\documentclass[review,10pt]{elsarticle}
\usepackage[top=4.3cm,bottom=4.3cm,left=4.8cm,right=4.8cm]{geometry}
\usepackage[T1]{fontenc}
\usepackage{newtxtext,newtxmath}
\usepackage{graphicx,booktabs,amsmath,xcolor,enumitem,rotating,array,placeins}
\usepackage[font=footnotesize,labelfont=bf,skip=6pt]{caption}
\usepackage[hidelinks]{hyperref}
\graphicspath{{figures/}}
\biboptions{sort&compress}
\journal{Pattern Recognition}

\newcommand{\BelowPolaris}{22}
\newcommand{\BelowSpica}{19}

\newcommand{\CompositeEpochGapHigh}{1.1}
\newcommand{\CompositeEpochGapLow}{0.5}

\newcommand{\CompositeRanked}{33}

\newcommand{\CopiesSpicaPolaris}{6.4}

\newcommand{\CrossfitPlaceboMax}{0.4}

\newcommand{\DenebEpochCheckpoints}{51}
\newcommand{\DenebEpochCompared}{5}
\newcommand{\DenebEpochGapHigh}{0.9}
\newcommand{\DenebEpochGapLow}{0.1}
\newcommand{\DenebEpochRestMax}{0.15}
\newcommand{\DenebEpochSentence}{ For DENEB the gap on duplicated samples also opens during the first epochs, but it stays below one point throughout: 0.1 after the first epoch and at most 0.9 later, while all other samples differ by at most \DenebEpochRestMax{} points (\ref{app:epochs}).}

\newcommand{\DenebPairAbsent}{$+$0.33}
\newcommand{\DenebPairAbsentCI}{[$-$0.10, $+$0.76]}

\newcommand{\DetectionSharePolaris}{70}

\newcommand{\DupNebulaPolaris}{50.2}
\newcommand{\DupPolarisNebula}{24.4}
\newcommand{\DupPolarisSpica}{43.3}
\newcommand{\DupRowsNebulaPolaris}{13,111}
\newcommand{\DupRowsPolarisNebula}{806}

\newcommand{\DupRowsSpicaSpica}{342}
\newcommand{\DupSpicaPolaris}{68.4}

\newcommand{\EpochGapFive}{10.4}

\newcommand{\EpochGapOne}{2.5}

\newcommand{\ExpertSpreadHigh}{5.0}

\newcommand{\JudgedPolaris}{31}
\newcommand{\JudgedSpica}{30}
\newcommand{\LearnedClearEarlier}{8}
\newcommand{\LearnedClearFamily}{1}

\newcommand{\LearnedRankBest}{6}
\newcommand{\LearnedRankWorst}{12}
\newcommand{\MlpNebulaNote}{}
\newcommand{\MlpNebulaPolaris}{3.2}

\newcommand{\MlpSpicaPolaris}{8.4}
\newcommand{\NothinkExpertHigh}{3.0}
\newcommand{\NothinkExpertLow}{2.4}
\newcommand{\NothinkRestMax}{0.55}

\newcommand{\PairShareSpicaPolaris}{32}
\newcommand{\PearlAboveDup}{$+$6.9 [$+$6.3, $+$7.6]}
\newcommand{\PearlAboveImage}{$+$2.6 [$+$1.0, $+$4.3]}

\newcommand{\PearlAboveImageRows}{958}
\newcommand{\PearlAboveUnseen}{$+$2.2 [$+$0.5, $+$3.9]}

\newcommand{\PearlCompositeRel}{$+$1.11 [$+$0.93, $+$1.29]}

\newcommand{\PearlDupFirst}{99.8}

\newcommand{\PearlMatchPairAbsent}{$+$7.19}
\newcommand{\PearlMatchPairAbsentCI}{[$+$6.09, $+$8.31]}

\newcommand{\PearlOtherHigh}{$+$1.1}
\newcommand{\PearlOtherLow}{$+$0.6}
\newcommand{\PearlPolarisCleaned}{57.61}
\newcommand{\PearlPolarisRank}{12}
\newcommand{\PearlPolarisReleased}{61.32}

\newcommand{\PearlRelAbsent}{$+$1.20}
\newcommand{\PearlRelAbsentCI}{[$+$0.16, $+$2.26]}

\newcommand{\PearlRelPair}{$+$6.45}
\newcommand{\PearlRelPairAbsent}{$+$5.24}
\newcommand{\PearlRelPairAbsentCI}{[$+$4.08, $+$6.42]}

\newcommand{\PolosNebulaDupGain}{$+$1.69 [$+$1.17, $+$2.22]}

\newcommand{\PolosSpicaDupGain}{$+$2.48 [$+$2.07, $+$2.90]}

\newcommand{\RemovedShareNebula}{15.6}

\newcommand{\RemovedShareSpica}{13.7}
\newcommand{\RidgeAbove}{8}
\newcommand{\RidgeBelow}{1}
\newcommand{\RidgeClipPolosHigh}{$+$0.2}
\newcommand{\RidgeClipPolosLow}{$-$0.6}
\newcommand{\RidgeComparisons}{18}
\newcommand{\RidgeEncoderGapHigh}{1.9}
\newcommand{\RidgeEncoderGapLow}{1.2}
\newcommand{\RidgeGapHigh}{2.1}
\newcommand{\RidgeGapLow}{0.3}

\newcommand{\RidgeRecentRank}{4}

\newcommand{\RigelFreePair}{$-$1.84 [$-$3.09, $-$0.61]}

\newcommand{\SameLabelNebulaPolaris}{99.9}

\newcommand{\SameLabelSpicaPolaris}{96.7}

\newcommand{\SpicaCommonRatings}{10,187}
\newcommand{\SpicaIds}{75,535}
\newcommand{\SpicaNebulaCommonPairs}{3,287}
\newcommand{\SpicaNoRefs}{148,810}
\newcommand{\SpicaOnlyLowest}{89}
\newcommand{\SpicaOnlyPairs}{3,148}
\newcommand{\SpicaOwnDup}{2.2}
\newcommand{\SpicaOwnDupUpper}{3.8}
\newcommand{\SpicaPictures}{61,571}
\newcommand{\SpicaReencoded}{98}
\newcommand{\SpicaReencodedRows}{681}
\newcommand{\SpicaRows}{333,397}
\newcommand{\SpicaTestInTrain}{1,289}
\newcommand{\SpicaTestInTrainShare}{39.2}
\newcommand{\SpicaTestPictures}{3,287}

\newcommand{\ValidationSpread}{0.5}

\ifdefined\WithoutAppendix
\makeatletter
\newcommand{\supplabel}[2]{\global\@namedef{r@#1}{{Appendix~#2}{}{}{}{}}}
\AtBeginDocument{\supplabel{app:repro}{A}\supplabel{app:matrix}{B}\supplabel{app:spica}{C}\supplabel{app:details}{D}%
  \supplabel{app:crossfit}{E}\supplabel{app:epochs}{F}\supplabel{app:rigel}{G}\supplabel{app:components}{H}%
  \supplabel{app:full}{I}\supplabel{app:cases}{J}}
\makeatother
\newcommand{\supplementnote}{ Appendices A--J are provided as Supplementary Material.}
\else
\newcommand{\supplementnote}{}
\fi
\newcommand{\tablefont}{\linespread{1}\normalsize\setlength{\tabcolsep}{4pt}}
\newcommand{\apptablefont}{\linespread{1}\footnotesize\setlength{\tabcolsep}{3.5pt}}
\newcommand{\smalltablefont}{\linespread{1}\scriptsize\setlength{\tabcolsep}{3pt}}
\newcommand{\finding}[1]{\smallskip\noindent\textbf{#1}}
\setlist{nosep,leftmargin=1.4em}
\begin{document}

\begin{frontmatter}

\title{Do learned image captioning metrics generalize? Auditing duplicated samples in human-judgment benchmarks}

\author[1]{Xiaochen Zhang}
\author[1]{Jin Li}
\author[2]{Chunlong Xia}
\author[1]{Longjiang Guo\corref{cor}}
\ead{longjiangguo@snnu.edu.cn}
\cortext[cor]{Corresponding author.}
\affiliation[1]{organization={School of Artificial Intelligence and Computer Science, Shaanxi Normal University},
  addressline={No. 620, West Chang'an Avenue, Chang'an District},
  city={Xi'an},
  postcode={710119},
  state={Shaanxi},
  country={China}}
\affiliation[2]{organization={Baidu Inc.},
  addressline={Baidu Campus, No. 10 Shangdi 10th Street, Haidian District},
  city={Beijing},
  postcode={100085},
  country={China}}

\begin{abstract}
Learned image captioning metrics are trained and tested on human judgments, so their reported correlation is a claim about generalization. We audit this claim for the three most recent collections of such judgments, Polaris, Nebula, and Spica, matching images at the pixel level. First, test samples reappear in training sets: the Spica training set contains about two-thirds of the Polaris test set, with identical human ratings. Second, controlled retraining shows that duplicates can inflate a metric: for Pearl, the gain is confined to the duplicated samples and grows with every epoch; for DENEB, it stays below one point. The gain depends on the learner and on how often a training set repeats a sample. Third, Kendall's $\tau$ over all samples is the sum of two terms, detecting captions with the lowest human score and ranking the others; the first supplies most of the reported correlation and is nearly saturated. The recent test sets cannot order Polos, DENEB and Pearl consistently, and a ridge regression on frozen features matches or exceeds all three. Fourth, we propose a protocol that ranks a learned metric only with a model trained without test samples, reports both terms, and compares metrics by paired tests. Under it, zero-shot vision--language judges rank first on the three recent test sets, and a fine-tuned judge ranks first on the three earlier benchmarks. The same analysis applies to any evaluator learned from human judgments; filter lists and code are released at \url{https://github.com/morninghut/ic-dataset-audit}.
\end{abstract}


\end{frontmatter}

\section{Introduction}

Automatic metrics decide which image captioning model is reported as better~\cite{stefanini2023survey,cao2024cast,li2025m3ixup}. Increasingly, the metrics are themselves learned: Polos~\cite{wada2024polos}, DENEB~\cite{matsuda2024deneb} and Pearl~\cite{hirano2026pearl} regress human scores from the features of pretrained encoders, and Rigel~\cite{koyama2026rigel} fine-tunes a vision--language model (VLM) for the same purpose. A learned metric is a pattern recognition model like any other. Its correlation with human scores on a test set is therefore a statement about generalization, and it deserves the scrutiny that the field applies to classifiers: is the test set independent of the training set, and what does the reported number measure?

This line of work became possible through three collections of human judgments, Polaris, Nebula and Spica, released with Polos, DENEB and Pearl. Together, they provide several hundred thousand ratings, far more than any earlier benchmark, and they are the reason learned metrics can be trained at all. The collections were built in sequence and draw on shared source images, and later papers also report earlier metrics on the newer test sets: Polos, for example, is a baseline on the Nebula test set in the papers of DENEB, Pearl and Rigel. If samples of a test set are also present in the training set of another collection, a metric evaluated across collections is partly tested on what it was trained on. To help the community use these resources reliably across generations, we examine whether this happens and what follows from it.

\begin{figure}[!t]
\centering
\includegraphics[width=\linewidth]{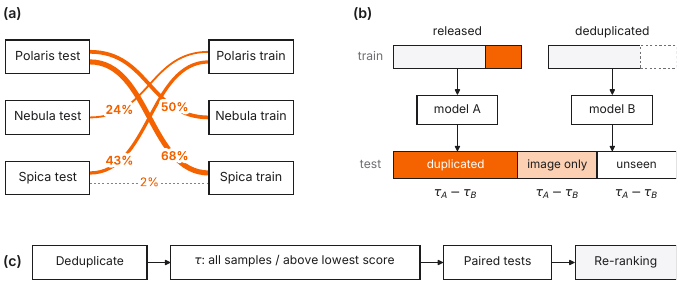}
\caption{Overview of the study. (a)~Audit. A line connects a test set to a training set that contains some of its samples (same image, same candidate caption, with a human score); the label and the line width give the share of the test set. (b)~Effect. The same metric is trained on the released training set (model A) and on the training set without the duplicated samples (model B). The test set is divided into samples that the released training set duplicates, samples of which it holds only the image, and samples whose image it has not seen. A gain $\tau_A-\tau_B$ that appears only in the first group points to memorization. (c)~Protocol. Learned metrics are ranked only with a deduplicated model, Kendall's $\tau$ is reported over all samples and over samples above the lowest human score, and metrics are compared by paired tests before \CompositeRanked{} metrics are re-ranked.}
\label{fig:overview}
\end{figure}

Fig.~\ref{fig:overview} summarizes the study in three steps, each of which answers one question.

\finding{Are test samples in the training sets?} Yes, across collections. About two-thirds of the Polaris test set is contained in the Spica training set and half of it in the Nebula training set; in the other direction, the Polaris training set contains large parts of the Spica and Nebula test sets. The duplicated samples carry the same human ratings on both sides. Within a collection and on the earlier benchmarks, Flickr8k and Composite, we find almost none.

\finding{Does this change the measured correlation?} It can, and how much depends on the metric. For Pearl, the gain from the duplicates sits on the duplicated samples and grows with every training epoch, while the other samples of the same test set stay flat. For DENEB, the gain is below one point, and for the released Rigel checkpoint we detect none. Cross-fitting three further learners shows the pattern behind this: the more a model can memorize, and the more often the training set repeats a sample, the more a duplicate is worth.

\finding{What do the benchmarks measure once duplicates are set aside?} A quarter or more of every evaluation set consists of captions with the lowest human score, most of them obvious mismatches. Kendall's $\tau$ over all samples decomposes exactly into a term for detecting these samples and a term for ranking the others. Detection is a necessary ability, but modern metrics have nearly saturated it, and it supplies most of the reported number. On the ranking term, the recent test sets do not separate Polos, DENEB and Pearl. \emph{A ridge regression on frozen features, the simplest model that can be fitted to human scores, matches or exceeds all three on each of these test sets.} Under a protocol that removes duplicates and reports both terms, zero-shot VLMs used as judges rank first on the recent test sets, and a fine-tuned judge on the earlier benchmarks.

\smallskip
\emph{Scope.} The audit concerns one family of collections, but the question is general: evaluators learned from human judgments are now common in machine translation, summarization, and image quality assessment, and their benchmarks also grow by extending earlier ones. We intend this study as a worked example of how the generalization of such an evaluator can be measured with statistical care, and as a contribution to the discussion in pattern recognition on what a benchmark number means~\cite{musgrave2020reality,kapoor2023leakage}.

We do not claim that published results are wrong. Each of the four papers evaluates its metric on sets that its own training set does not duplicate, and we reproduce their numbers (\ref{app:repro}).\supplementnote{} What duplication affects is the comparison \emph{across} collections. Our contributions are:
\begin{itemize}
\item a content-level audit of three training sets and six evaluation sets, with exact and near-duplicate matching and worked examples (Section~\ref{sec:audit});
\item an estimate of the inflation caused by duplicated samples from four versions of each training set, three seeds, cross-fitting of three further learners and per-epoch curves (Section~\ref{sec:inflation});
\item an exact decomposition of Kendall's $\tau$ into a detection and a ranking term, with paired tests and a human reference that show what each benchmark can resolve (Section~\ref{sec:validity});
\item a protocol built on these results, a re-ranking of \CompositeRanked{} metrics that includes a linear baseline and zero-shot judges, case studies, and released filter lists and code (Section~\ref{sec:protocol}).
\end{itemize}

\section{Related work}
\label{sec:related}

\paragraph{Metrics for image captioning} Two recent surveys organize more than seventy metrics~\cite{berger2025surveying,sarto2025survey}; we follow their grouping into four families. \emph{Text-overlap} metrics compare a candidate with human references by n-grams or scene graphs: BLEU~\cite{papineni2002bleu}, METEOR~\cite{banerjee2005meteor}, ROUGE~\cite{lin2004rouge}, CIDEr~\cite{vedantam2015cider} and SPICE~\cite{anderson2016spice}. \emph{Embedding-similarity} metrics compare the candidate with the image in the space of a contrastive encoder~\cite{radford2021clip}. CLIPScore~\cite{hessel2021clipscore} introduced the idea and needs no references; PAC-S~\cite{sarto2023pacs} and PAC-S++~\cite{sarto2025pacspp} refine the encoder with positive-augmented contrastive learning and remain training-free with respect to human scores. \emph{Learned regression} metrics fit human scores directly: Polos, DENEB and Pearl. \emph{Judges} prompt a large model for a rating: CLAIR~\cite{chan2023clair} and G-VEval~\cite{tong2024gveval} call a proprietary model, FLEUR~\cite{lee2024fleur} uses an open VLM and reads the score from token probabilities, and Rigel fine-tunes such a judge on human scores. EXPERT~\cite{kim2025expert} is closest to our topic among the metric papers: its authors noticed that Polaris and Nebula share image--caption pairs and merged them when building a combined training set. They did not examine whether the shared pairs cross the boundary between training and test splits, which is the question we pursue.

\paragraph{How metrics are evaluated} Metrics are compared by their correlation with human judgments on Flickr8k~\cite{hodosh2013flickr} and Composite~\cite{aditya2015composite}, and since 2024, on the test splits of Polaris and Nebula. An earlier re-evaluation asked what these correlations capture~\cite{kilickaya2017re}. Neighbouring fields have gone further on the statistical side. In machine translation, paired resampling became the standard test~\cite{koehn2004significance}, and the treatment of ties in Kendall's $\tau$ was shown to change which metric wins~\cite{deutsch2023ties}. For classifiers, Dem\v{s}ar~\cite{demsar2006statistical} established that comparisons over several data sets need tests corrected for multiple comparisons. In deep metric learning, Musgrave et al.~\cite{musgrave2020reality} found that years of reported progress shrank once methods were compared under one protocol. Published tables of captioning metrics still report a single correlation per benchmark; we bring these tools to this setting.

\paragraph{Duplicates, leakage and contamination} Near-duplicate test images in CIFAR inflate accuracy~\cite{barz2020cifar}, and rebuilt ImageNet test sets lower it~\cite{recht2019imagenet}. In open-domain question answering, most test answers also appear in the training data~\cite{lewis2021overlap}. Leakage is a recurring cause of irreproducible results across the sciences that use machine learning~\cite{kapoor2023leakage}, and large pretrained models may have seen entire benchmarks~\cite{sainz2023contamination}, images included~\cite{ramos2025leakage}. Methods to detect such overlap range from hashing to retrieval in a feature space~\cite{barz2020cifar,ramos2025leakage}; we use both. Networks can memorize individual labels, and the way to measure it is to train with and without an example and compare predictions on it~\cite{feldman2020memorize}; our cross-fitting design applies this idea to duplicated evaluation samples. Our case differs from these studies in three ways. What is duplicated is not an input but a \emph{labelled evaluation sample}, an image with a candidate caption and its human score, and the affected model is itself an evaluator. The duplication happens between sibling collections that are evaluated on each other, not between a web-scale corpus and a benchmark. And we do not stop at counting overlap but estimate its effect with controlled retraining, a matched control and a per-epoch dose--response curve. To the best of our knowledge, this is the first train--test audit of the human-judgment benchmarks on which learned captioning metrics are developed.

\section{Benchmarks, terms and statistical tools}
\label{sec:background}

\subsection{Collections and evaluation sets}

Polaris~\cite{wada2024polos}, Nebula~\cite{matsuda2024deneb} and Spica~\cite{hirano2026pearl} each provide training, validation and test splits. Every entry consists of an image, human-written reference captions, a candidate caption, typically the output of a captioning model, and a human score on a five-level scale. Polos is trained on the Polaris training set, DENEB on that of Nebula, while Pearl and Rigel are trained on that of Spica. We evaluate on the three test sets and on the earlier benchmarks Flickr8k-Expert, Flickr8k-CF~\cite{hodosh2013flickr} and Composite~\cite{aditya2015composite}.

The collections store judgments in two different ways, and the difference is needed to read every count in this paper (Table~\ref{tab:structure}). Polaris and Spica store \emph{one entry per rating}: an image--caption pair that was rated by eight annotators appears eight times, each time with one score. Nebula stores \emph{one entry per pair}, with the mean of its ratings. The Polaris test set therefore has about three times as many entries as image--caption pairs. Two further facts follow from the table. First, about two-thirds of the pairs in Polaris (63\% in the training set and in the test set) carry a single rating, and that rating is almost always the lowest score (99.7\%); the pairs that were rated several times make up the rest. Table~\ref{tab:groups} shows where these pairs fall in the test set. Second, the Spica test set and the Nebula test set are largely one set of judgments in two forms: the Spica test set contains the individual ratings of the Nebula test pairs, and further pairs that mostly carry a single lowest score (\ref{app:spica}). Results on these two test sets are thus not independent of each other.

\begin{table}[!t]
\caption{How each set stores human judgments. A pair is an image with one candidate caption; a sample is one stored entry, which is one rating in Polaris, Spica, and Flickr8k-Expert, and one pair with its mean score elsewhere. ``Lowest'' is the share of samples with the lowest score on the scale. Indented lines give the training sets after deduplication (Section~\ref{sec:inflation}). Images are counted by pixel content.}
\label{tab:structure}
\centering\tablefont
\begin{tabular}{@{}llrrrrr@{}}
\toprule
& A sample & & & & Ratings & Lowest \\
Set & is one & Images & Pairs & Samples & per pair & (\%) \\
\midrule
Polaris train & rating & 8,214 & 26,340 & 78,630 & 3.0 & 24 \\
\quad deduplicated & & 7,984 & 24,203 & 65,051 & 2.7 & 28 \\
Polaris test & rating & 2,738 & 8,726 & 26,122 & 3.0 & 24 \\
Nebula train & pair & 26,382 & 26,382 & 26,382 & 1.0 & 27 \\
\quad deduplicated & & 22,271 & 22,271 & 22,271 & 1.0 & 32 \\
Nebula test & pair & 3,298 & 3,298 & 3,298 & 1.0 & 26 \\
Spica train & rating & 57,999 & 106,532 & 309,386 & 2.9 & 15 \\
\quad deduplicated & & 56,396 & 98,263 & 266,893 & 2.7 & 16 \\
Spica test & rating & 3,287 & 6,435 & 15,238 & 2.4 & 26 \\
\addlinespace
Flickr8k-Expert & rating & 1,000 & 5,664 & 16,992 & 3.0 & 58 \\
Flickr8k-CF & pair & 1,000 & 47,830 & 47,830 & 1.0 & 88 \\
Composite & pair & 3,964 & 11,962 & 11,982 & 1.0 & 28 \\
\bottomrule
\end{tabular}

\end{table}

\subsection{Terms}

Table~\ref{tab:terms} fixes our terms. We call the unit over which a correlation is computed a \emph{sample}, however the set stores it, and count in samples unless stated otherwise.

\begin{table}[!t]
\caption{Terms used in this paper.}
\label{tab:terms}
\centering\tablefont
\begin{tabular}{@{}>{\raggedright\arraybackslash}p{3.3cm}p{7.8cm}@{}}
\toprule
Term & Meaning \\
\midrule
Candidate & The caption to be scored. \\
Reference & A caption written by a person for the same image. \\
Pair & An image together with one candidate. \\
Rating & The score that one annotator gave to a pair. \\
Sample & One stored entry of a set: a rating, or a pair with its mean score (Table~\ref{tab:structure}). Correlations are computed over samples. \\
Duplicated sample & An evaluation sample whose pair also occurs, with a human score, in a given training set. \\
Image-only sample & An evaluation sample whose image occurs in the training set, but only with other candidates. \\
Unseen-image sample & An evaluation sample whose image is not in the training set. \\
Released, deduplicated training set & The training set as published / after removing every sample whose pair occurs in any evaluation set. \\
Lowest-score sample & A sample that carries the lowest score of its scale. \\
Inflation & The part of a metric's correlation on an evaluation set that it owes to having been trained on samples of that set. \\
Human reference & The score obtained when a held-out human rating takes the place of the metric. \\
\bottomrule
\end{tabular}
\end{table}

\subsection{Statistical tools}
\label{sec:tools}

The comparisons below rest on four standard tools, which we briefly introduce here.

\emph{Kendall's $\tau$.} A metric is good if it orders captions as people do. Kendall's $\tau$~\cite{kendall1938} takes every pair of samples that humans scored differently and checks whether the metric orders them the same way (concordant) or the opposite way (discordant). With $C$ concordant and $D$ discordant pairs among $n$ samples whose human scores take $m$ distinct values,
\begin{equation}
\tau_c = \frac{2\,(C-D)}{n^{2}\,(m-1)/m}.
\label{eq:tau}
\end{equation}
We follow the metric papers in reporting $100\times\tau_c$ and $\tau_b$ on Flickr8k-CF; we write $\tau$ for both and call one unit a point.

\emph{Bootstrap over images.} A test set is one draw from all the images that could have been collected, and a different draw would give a different $\tau$. The bootstrap~\cite{efron1993bootstrap} imitates this by drawing images from the test set at random, with replacement, many times, and recomputing $\tau$ each time; the middle 95\% of the results is the interval we report. We draw whole images, with all their samples, because samples of one image are not independent.

\emph{Paired comparison.} Two metrics evaluated on the same samples share most of their luck: an easy draw of images helps both, and comparing their separate intervals ignores this. We compute the \emph{difference} between the two metrics on each bootstrap draw and report the interval of that difference~\cite{koehn2004significance}. A difference whose interval excludes zero is one that the test set can resolve.

\emph{Correction for multiple comparisons.} Among the several hundred pairs that can be formed from more than thirty metrics, some differences look significant by chance alone. Holm's procedure~\cite{holm1979} raises the bar for each test according to how many tests are made, so that the chance of even one false claim stays below 5\%. All statements that two metrics are ``separated'' refer to this corrected test.

Learned metrics are trained three times with different random seeds, and we report the mean; a single training can differ from another by more than the differences between metrics (Section~\ref{sec:protocol}).

\section{Audit: are there test samples in the training sets?}
\label{sec:audit}

\subsection{Matching by content}

File names and identifiers are not stable across the collections, so we match content. Two images are the same if their decoded pixels have the same hash; two candidates are the same if their text is identical (details in \ref{app:details}). For a training set $T$ and an evaluation set, every evaluation sample $s$ with image $x_s$ and candidate $c_s$ then falls into one of three groups,
\begin{equation}
g_T(s)=\begin{cases}
\text{duplicated} & \text{if } (x_s,c_s) \text{ occurs in } T,\\
\text{image only} & \text{if } x_s \text{ occurs in } T \text{, but not with } c_s,\\
\text{unseen image} & \text{otherwise.}
\end{cases}
\label{eq:groups}
\end{equation}

\subsection{What is duplicated}

\begin{figure}[!t]
\centering
\includegraphics[width=\linewidth]{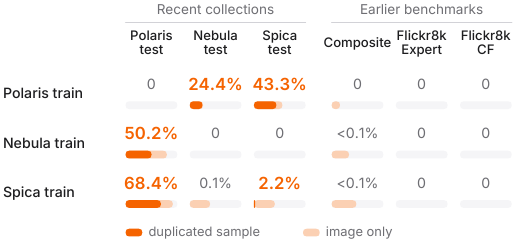}
\caption{Share of each evaluation set that is inside each training set. The number and the dark bar indicate the duplicated samples; the light bar indicates the image-only samples. Training sets: Polaris (Polos), Nebula (DENEB), and Spica (Pearl and Rigel).}
\label{fig:duplication}
\end{figure}

\begin{figure}[!t]
\centering
\includegraphics[width=\linewidth]{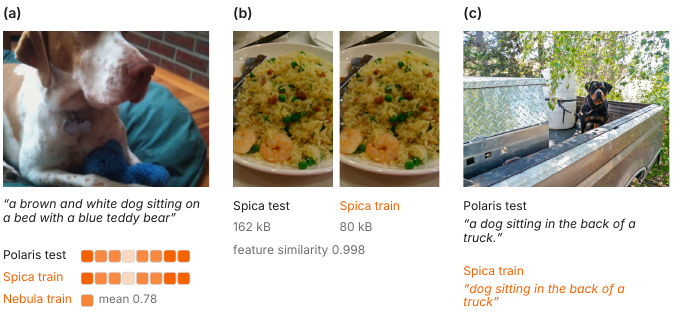}
\caption{Three kinds of duplicates, each with a real example. (a)~Exact duplicate: the same image file and candidate appear in the Polaris test set and in the Spica training set with the same eight ratings (one square per rating, darker is higher), and in the Nebula training set with the same mean. (b)~Re-encoded image: one photograph stored as two files of equal size in pixels whose hashes differ; it is found only by feature similarity. (c)~Near-identical caption: the same image, a candidate that differs by an article and a period.}
\label{fig:examples}
\end{figure}

\emph{Samples move between collections.} Fig.~\ref{fig:duplication} gives the result for every training set and evaluation set. The Spica training set contains \DupSpicaPolaris\% of the samples of the Polaris test set, and the Nebula training set contains \DupNebulaPolaris\%. In the other direction, the Polaris training set contains \DupPolarisSpica\% of the Spica test set and \DupPolarisNebula\% of the Nebula test set. Within a collection, the overlap is small (\SpicaOwnDup\% for Spica, none for the others).

\emph{The duplicates carry the same ratings.} Fig.~\ref{fig:examples}a shows a typical case: the eight ratings of a pair appear in the same order in the Polaris test set and in the Spica training set, and the Nebula training set stores their mean. This is the rule. After averaging the ratings of each pair on both sides, the two copies agree exactly for \SameLabelSpicaPolaris\% of the duplicated pairs between the Spica training set and the Polaris test set, and for \SameLabelNebulaPolaris\% between the Nebula training set and the Polaris test set (\ref{app:matrix}). A metric trained on either set is thus shown the test labels themselves, not a second opinion on the same pair.

\emph{Counting pairs instead of samples.} Because the collections store judgments differently, the same overlap looks smaller when counted in pairs: the duplicates in the Spica training set are \PairShareSpicaPolaris\% of the Polaris test pairs. These are the pairs that were rated several times, so they hold \DupSpicaPolaris\% of the ratings, and it is the ratings over which $\tau$ is computed. Most of the remaining pairs carry a single lowest score.

\emph{Images are shared more widely than judgments.} Almost half of the Composite samples show an image that is also in the Spica training set, but only one is a duplicated sample. Flickr8k-Expert and Flickr8k-CF share nothing with any training set.

\subsection{Exact matching misses little}

Hashing cannot find a photograph that was saved twice (Fig.~\ref{fig:examples}b) or a caption that differs by a character (Fig.~\ref{fig:examples}c). We therefore repeat the audit with a relaxed rule: images match if the cosine similarity of their SigLIP~\cite{zhai2023siglip} features exceeds 0.97, and candidates match if they are identical after lower-casing and removing punctuation or have a character similarity above 0.9. The four large shares rise by less than one percentage point (\ref{app:matrix}). The exact rule is thus a tight lower bound, and all experiments use it. The released filter lists come in both versions; the relaxed one removes slightly more training samples, 45{,}371 instead of 42{,}493 from the Spica training set.

\FloatBarrier
\section{Do duplicated samples inflate a metric?}
\label{sec:inflation}

\subsection{Design: compare within the test set}

The obvious experiment is to retrain a metric without the duplicates and see whether its score drops. A drop alone proves little, because removing data can weaken a model for reasons unrelated to memorization. We therefore use a \emph{difference-in-differences}: the same two models are compared on the duplicated samples and on the other samples of the same test set. A general loss of quality lowers both groups; memorization lowers only the first. Formally, let $\tau_g(M)$ be Kendall's $\tau$ of model $M$ on the samples of group $g$ in Eq.~\eqref{eq:groups}, let $M_{\text{rel}}$ and $M_{\text{ded}}$ be the models trained on the released and on the deduplicated training set, and
\begin{equation}
\Delta_g=\tau_g(M_{\text{rel}})-\tau_g(M_{\text{ded}}),\qquad
I=\Delta_{\text{duplicated}}-\Delta_{\text{unseen image}}.
\label{eq:did}
\end{equation}
$\Delta_g$ is the gain of the released training set within one group, and $I$ is the part of that gain that is specific to duplicated samples.

We retrain Pearl and DENEB from their public code and settings, three seeds each, on four versions of their training set:
\begin{itemize}
\item \emph{released}: as published;
\item \emph{deduplicated}: without every sample whose pair occurs in any evaluation set (\RemovedShareSpica\% of the samples of Pearl's training set and \RemovedShareNebula\% of DENEB's; Table~\ref{tab:structure});
\item \emph{random removal}: without the same number of samples chosen at random;
\item \emph{matched removal} (Pearl only): without the same number of samples that resemble the removed ones (with references, same number of ratings, same score range) but are duplicated nowhere, so that the duplicates stay.
\end{itemize}
The last two are controls for the size of the training set. All differences are means over three seeds with 95\% bootstrap intervals over images.

\subsection{Retraining: the gain sits on the duplicated samples}

\begin{figure}[!t]
\centering
\includegraphics[width=\linewidth]{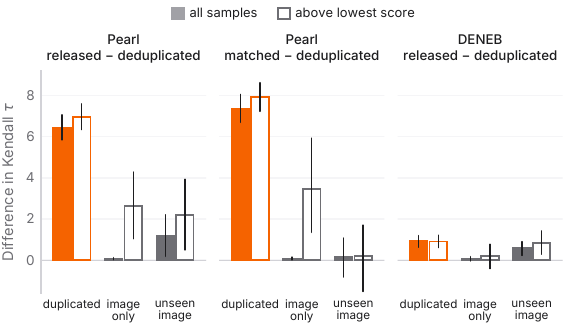}
\caption{Gain in Kendall's $\tau$ on the Polaris test set from training on a set that holds the duplicates, relative to the deduplicated training set, within each group of test samples (mean of three seeds, 95\% interval). Filled bars: all samples of the group. Open bars: only samples above the lowest human score. ``Matched'' is the control that removes as much similar data as the deduplication but keeps the duplicates.}
\label{fig:groups}
\end{figure}

\emph{Pearl.} Trained on the released set, Pearl reaches \PearlPolarisReleased{} on the Polaris test set; trained on the deduplicated set, \PearlPolarisCleaned. Fig.~\ref{fig:groups} locates the difference: about six and a half points on the duplicated samples and about one on unseen images, so the specific gain $I$ of Eq.~\eqref{eq:did} is \PearlRelPairAbsent{} \PearlRelPairAbsentCI. The matched control settles what the gain is made of. It gives up as much similar data as the deduplication does, yet it keeps the full gain on duplicated samples and has none on unseen images ($I=$ \PearlMatchPairAbsent{} \PearlMatchPairAbsentCI). The gain therefore comes from the duplicated samples and not from the size of the training set.

\emph{DENEB.} The released training set is slightly better on duplicated samples and on unseen images alike. The specific gain, \DenebPairAbsent{} \DenebPairAbsentCI, includes zero, and the released set is better by a similar margin on evaluation sets with which it shares nothing. We find no significant inflation specific to duplicates and bound it below one point; the per-epoch curves of Section~\ref{sec:dose} agree with an effect of at most that size.

\subsection{Image-only samples are uninformative}
\label{sec:imageonly}

\begin{table}[!t]
\caption{The three groups of Polaris test samples relative to the Spica training set. ``Single'' is the share of samples whose pair has one rating; ``tied'' is the share of sample pairs with equal human scores, which cannot contribute to $\tau$. $\Delta$ is the gain of Pearl from the released training set, Eq.~\eqref{eq:did}, over all samples of the group and over samples above the lowest score.}
\label{tab:groups}
\centering\tablefont
\begin{tabular}{@{}lrrrrrr@{}}
\toprule
& & Lowest & Single & Tied & \multicolumn{2}{c}{$\Delta$ (Pearl)} \\
\cmidrule(l){6-7}
Group & Samples & (\%) & (\%) & (\%) & all & above \\
\midrule
Duplicated & 17,856 & 3 & 0 & 29 & $+$6.4 & $+$6.9 \\
Image only & 5,987 & 84 & 84 & 71 & $+$0.1 & $+$2.6 \\
Unseen image & 2,279 & 24 & 22 & 23 & $+$1.2 & $+$2.2 \\
\bottomrule
\end{tabular}

\end{table}

In Fig.~\ref{fig:groups}, the filled bar of the image-only samples is almost exactly zero, with a very narrow interval. This does not show that sharing an image is harmless. Table~\ref{tab:groups} explains the pattern: the image-only group consists mostly of lowest-score samples whose pairs were rated once, so most of its sample pairs are tied in the human score, and the rest are easy for every model. Kendall's $\tau$ on such a group barely reacts to any change in the model.

The open bars in Fig.~\ref{fig:groups} therefore repeat the comparison on samples above the lowest score. For Pearl, the gain of the released set is then \PearlAboveImage{} on image-only samples and \PearlAboveUnseen{} on unseen images, against \PearlAboveDup{} on duplicated samples. The duplicated group still stands out by more than four points. The image-only group does not differ from the unseen group, but it is small above the lowest score (\PearlAboveImageRows{} samples), and its interval is wide. We conclude that there is no evidence for an effect of shared images alone, and that these test sets cannot exclude one of a few points.

\subsection{Cross-fitting: the effect depends on the learner}
\label{sec:crossfit}

Retraining is expensive and removes data. \emph{Cross-fitting}~\cite{chernozhukov2018double} measures the same effect without removing anything, and for any learner. The images of a training set are divided into five folds, and five models are trained, each without one fold. For every sample, one model has not seen its image and four have. Because the five models are trained in the same way, any difference between their scores on a sample reflects whether the sample was seen~\cite{feldman2020memorize}. Let $u(s)$ be the score of the model that did not see sample $s$ and $v_j(s)$, $j=1,\dots,4$, the scores of the four that did. Within a group $g$,
\begin{equation}
E_g=\frac{1}{4}\sum_{j=1}^{4}\tau_g(v_j)-\tau_g(u).
\label{eq:crossfit}
\end{equation}
Each term compares one model with one model; the four seen models are never averaged into an ensemble, which would raise $\tau$ by itself.

\begin{figure}[!t]
\centering
\includegraphics[width=\linewidth]{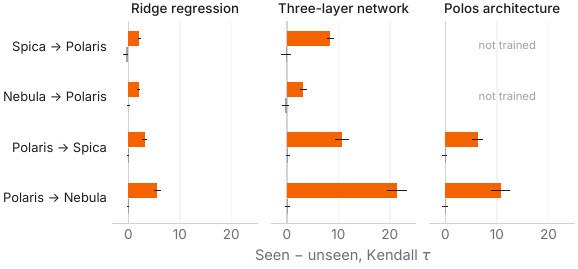}
\caption{Cross-fit effect $E_g$ of having trained on a sample, Eq.~\eqref{eq:crossfit}, for three learners of increasing capacity (networks are means of three seeds). Each row is a training set, and the test set shares samples with it. Orange bars: duplicated samples. Grey bars: samples whose image is unseen, a placebo that should show no effect. Whiskers: 95\% intervals.}
\label{fig:crossfit}
\end{figure}

\begin{table}[!t]
\caption{Cross-fit effect on duplicated samples: $\tau$ of the model that did not see a sample and mean $\tau$ of the four that did, their difference $E$ (95\% interval), and the range of the four single-model differences.}
\label{tab:crossfit}
\centering\tablefont
\begin{tabular}{@{}lrrrr@{}}
\toprule
Learner, training $\rightarrow$ test & Unseen & Seen & $E$ [95\% CI] & Single models \\
\midrule
\multicolumn{5}{@{}l}{\emph{Ridge regression, SigLIP features}} \\
\quad Spica $\rightarrow$ Polaris & 28.9 & 31.1 & $+$2.1 [$+$2.0, $+$2.4] & $+$2.1 to $+$2.2 \\
\quad Nebula $\rightarrow$ Polaris & 29.6 & 31.6 & $+$2.0 [$+$1.7, $+$2.3] & $+$1.9 to $+$2.1 \\
\quad Polaris $\rightarrow$ Spica & 31.4 & 34.6 & $+$3.2 [$+$2.7, $+$3.7] & $+$3.0 to $+$3.3 \\
\quad Polaris $\rightarrow$ Nebula & 43.0 & 48.6 & $+$5.6 [$+$4.9, $+$6.3] & $+$5.1 to $+$5.9 \\
\multicolumn{5}{@{}l}{\emph{Three-layer network, SigLIP features}} \\
\quad Spica $\rightarrow$ Polaris & 28.0 & 36.4 & $+$8.4 [$+$7.7, $+$9.1] & $+$6.9 to $+$9.6 \\
\quad Nebula $\rightarrow$ Polaris & 27.4 & 30.6 & $+$3.2 [$+$2.5, $+$3.8] & $+$2.7 to $+$4.1 \\
\quad Polaris $\rightarrow$ Spica & 30.0 & 40.7 & $+$10.7 [$+$9.4, $+$12.1] & $+$8.8 to $+$12.2 \\
\quad Polaris $\rightarrow$ Nebula & 41.6 & 63.0 & $+$21.4 [$+$19.4, $+$23.4] & $+$16.9 to $+$24.7 \\
\multicolumn{5}{@{}l}{\emph{Polos architecture, CLIP features}} \\
\quad Polaris $\rightarrow$ Spica & 27.6 & 33.9 & $+$6.3 [$+$5.2, $+$7.3] & $+$5.4 to $+$7.1 \\
\quad Polaris $\rightarrow$ Nebula & 37.8 & 48.6 & $+$10.9 [$+$8.9, $+$12.6] & $+$9.6 to $+$12.3 \\
\bottomrule
\end{tabular}

\end{table}

We apply this to three learners on the same frozen features: a ridge regression, a three-layer network, and the architecture of Polos (Fig.~\ref{fig:crossfit}, Table~\ref{tab:crossfit}; settings in \ref{app:details}, all groups in \ref{app:crossfit}). All three gain on duplicated samples, while the placebo group of unseen images stays within \CrossfitPlaceboMax{} points of zero. The size of the gain follows the capacity of the learner: a few points for the linear model and several times as much for the networks. The same duplicates are thus worth very different amounts to different models, which is why the audit of Section~\ref{sec:audit} alone cannot say how much a published number is inflated.

\subsection{Dose--response: the gain grows with every epoch}
\label{sec:dose}

\begin{figure}[!t]
\centering
\includegraphics[width=\linewidth]{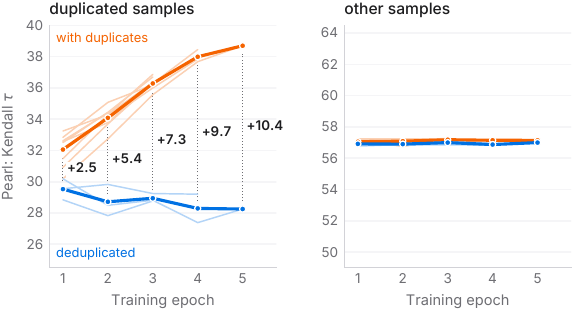}
\caption{Kendall's $\tau$ on the Polaris test set after each training epoch, on duplicated samples (left) and on all other samples (right). Thin lines are single Pearl trainings; thick lines are the mean of the trainings whose training set holds the duplicates (orange) or is deduplicated (blue); numbers are the gap between the two means. Trainings stop at different epochs, so later means average fewer runs. Both panels span the same number of points. DENEB: \ref{app:epochs}.}
\label{fig:epochs}
\end{figure}

If a gain is memorization, more exposure should produce more of it. This is the logic of a \emph{dose--response} relation: an effect that rises with the dose is hard to attribute to anything else. Here the dose is the number of passes over the training set. The public scripts store a checkpoint after every epoch; we scored them all (Fig.~\ref{fig:epochs}).

For Pearl, the trainings whose set holds the duplicates climb by about two points per epoch on the duplicated samples, while the deduplicated trainings stay level; the gap widens from \EpochGapOne{} to \EpochGapFive{} points. On all other samples, every checkpoint of every training lies within half a point. This comparison follows each model against itself one epoch earlier, so it does not depend on whether the deduplicated set is in some way a weaker training set. It also explains why single trainings disagree: the validation score that selects the checkpoint varies by at most \ValidationSpread{} points between the epochs of a run, so the selected epoch, and with it the inflation that one training reports, is close to arbitrary.
\DenebEpochSentence

\subsection{Why Pearl gains more than DENEB}

The label audit rules out one explanation: both training sets hold the test ratings themselves (Section~\ref{sec:audit}), so DENEB is not protected by seeing different labels. Two differences remain, and the evidence supports both. \emph{Repetition.} The Spica training set stores a duplicated pair once per rating, on average \CopiesSpicaPolaris{} times, while the Nebula training set stores it once. Pearl thus meets each duplicated pair several times per epoch. \emph{The learner.} Cross-fitting holds the learner fixed and varies the training set: the same three-layer network gains \MlpSpicaPolaris{} points from the duplicates in the Spica training set and \MlpNebulaPolaris{} from those in the Nebula training set. The per-epoch curves agree: DENEB shows the signature of Pearl at about a tenth of the size (\DenebEpochGapHigh{} against \EpochGapFive{} points). The inflation is therefore a joint property of how a collection stores its judgments and of the model trained on it. A split of the duplicates by whether the two copies of the label agree, which would test a third explanation, is not possible: fewer than 4\% differ.

\subsection{Further checks}
\label{sec:checks}

\emph{The general cost of deduplication does not explain the gain.} A model trained on the deduplicated set is slightly weaker everywhere, also where nothing is duplicated: the released model is ahead by \PearlCompositeRel{} points on Composite, by \PearlRelAbsent{} \PearlRelAbsentCI{} on the unseen-image samples of the Polaris test set, and by \PearlOtherLow{} to \PearlOtherHigh{} over all samples of the other evaluation sets. This general cost of about one point is the same on Composite and on unseen images, and it is exactly the term that Eq.~\eqref{eq:did} subtracts. Its likely source is that deduplication deletes every judgment that originates from the Polaris validation and test splits, which the matched control keeps; for the estimate of inflation, the source does not matter, because the gain on duplicated samples differs from the general cost in three ways that a loss of useful training data cannot produce. \emph{Size.} It is \PearlRelPair{} points against about one, and the difference $I$ has an interval far from zero. \emph{Place.} It appears only on the test samples that the training set contains, and the matched control, which gives up as much data, keeps all of it. \emph{Time.} The general cost does not grow with training: on Composite, the gap stays between \CompositeEpochGapLow{} and \CompositeEpochGapHigh{} points at every epoch (\ref{app:epochs}). The gap on duplicated samples rises with every pass over the training set, from \EpochGapOne{} to \EpochGapFive{} points. A model that merely lacks useful data does not fall further behind, pass after pass, on exactly those test samples that its competitor has seen; a model that memorizes them does. The gain $I$ is therefore consistent with memorization of the duplicated samples; because the comparison is not a randomized experiment, it supports this reading rather than proving it.

\emph{Rigel.} Only the inference code and one checkpoint of Rigel are public, so it cannot be retrained. Compared group by group with a zero-shot judge that has no training set, its lead is not significantly larger on duplicated samples than on unseen images (\ref{app:rigel}). The same indirect comparison recovers only about half of the known effect for Pearl, so it cannot exclude an inflation of a few points. The exposure is also small: according to its paper, Rigel is fitted to human scores for a single epoch, and after one epoch the gap of Pearl is \EpochGapOne{} points.

\emph{Published numbers.} For the metric that each of the four papers proposes, the training set contains at most 0.1\% of any test set reported in that paper; these numbers are not in question. Duplicates enter published tables in two other ways. First, Polos is a baseline on the Nebula test set in the DENEB, Pearl and Rigel papers, although its training set contains \DupPolarisNebula\% of that set; trained on the released instead of the deduplicated Polaris set, the Polos architecture gains \PolosNebulaDupGain{} points there and \PolosSpicaDupGain{} on the Spica test set. Second, later metrics are trained on several collections at once. EXPERT~\cite{kim2025expert} is trained on the Polaris and Nebula training sets and reported on both test sets; its released training files contain \DupRowsNebulaPolaris{} of the 26{,}122 samples of the Polaris test set (\DupNebulaPolaris\%) and \DupRowsPolarisNebula{} of the 3{,}298 samples of the Nebula test set (\DupPolarisNebula\%), matched by image identifier and caption. BEiTScore~\cite{gomes2026beitscore} likewise trains on both training sets and reports both test sets. We have not retrained either and do not claim that their scores are inflated: DENEB shows that duplicates need not inflate a metric; Pearl shows that they can.

\FloatBarrier
\section{What do the benchmarks measure?}
\label{sec:validity}

\subsection{Kendall's \texorpdfstring{$\tau$}{tau} is a sum of detection and ranking}
\label{sec:components}

Every evaluation set contains many samples with the lowest score on its scale, from about a quarter of the recent test sets to almost nine-tenths of Flickr8k-CF (Table~\ref{tab:structure}). Most are captions that do not belong to the image. Telling them apart from acceptable captions is a real and necessary ability of a metric. It is a different ability from ordering acceptable captions by quality, and Kendall's $\tau$ adds the two without saying so.

\begin{figure}[!t]
\centering
\includegraphics[width=\linewidth]{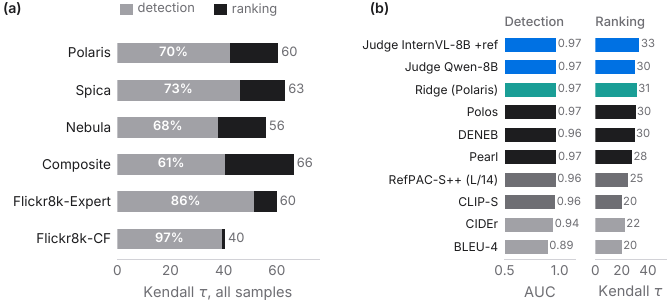}
\caption{The two terms of Eq.~\eqref{eq:components}. (a)~Kendall's $\tau$ over all samples of the best-ranked metric of each evaluation set, divided into the detection term (grey, with its share) and the ranking term (black). (b)~Ten ranked metrics on the Polaris test set: detection ability (AUC for separating lowest-score samples from the others; bars start at chance, 0.5) and ranking ability ($\tau$ above the lowest score).}
\label{fig:components}
\end{figure}

The split is exact. Let $L$ be the lowest-score samples and $A$ the samples above the lowest score, with $n_L$ and $n_A$ members. Two samples of $L$ are tied in the human score and do not count. A sample of $L$ and a sample of $A$ are ordered correctly by the metric with a probability that is, by definition, the area under the ROC curve (AUC) of the metric for separating $A$ from $L$. Counting concordant minus discordant pairs in Eq.~\eqref{eq:tau} separately for the two kinds of pairs gives
\begin{equation}
\tau_c(\text{all})=\underbrace{w_{\text{det}}\,(2\,\mathrm{AUC}-1)}_{\text{detection}}+\underbrace{w_{\text{rank}}\;\tau_c(A)}_{\text{ranking}},
\label{eq:components}
\end{equation}
\begin{equation}
w_{\text{det}}=\frac{2m}{m-1}\,\frac{n_L\,n_A}{n^{2}},\qquad
w_{\text{rank}}=\frac{m\,(m_A-1)}{(m-1)\,m_A}\,\frac{n_A^{2}}{n^{2}},
\label{eq:weights}
\end{equation}
where $m$ and $m_A$ are the numbers of distinct human scores in all samples and in $A$. The weights depend only on the composition of the evaluation set, not on the metric.

Fig.~\ref{fig:components}a applies Eq.~\eqref{eq:components} to the best metric of each set (all sets and metrics: \ref{app:components}). Detection supplies about \DetectionSharePolaris\% of its $\tau$ on the Polaris test set and nearly all of it on Flickr8k-CF. Fig.~\ref{fig:components}b shows why this matters for comparing metrics. On detection, the metrics of the last years lie within a narrow band at an AUC of about 0.96; the ability is close to saturated. A simple check confirms how easy the task is: on the Polaris and Spica test sets, the cosine similarity between the CLIP text embeddings~\cite{radford2021clip} of the candidate and of its closest reference, computed without looking at the image, separates the samples whose only rating is the lowest score from all others with an AUC of 0.98. On ranking, the same metrics spread over more than ten points of $\tau$. A table that reports only $\tau$ over all samples thus compresses the differences that remain between metrics into a small part of a number dominated by an ability they share.

Reporting both terms changes the order of metrics. On the three recent test sets, CLIP-S is above BLEU-1 on all samples and below it above the lowest score. A likely reason is that BLEU-1 can use the words of the human references as anchors to separate small differences in quality, whereas reference-free embedding similarity was designed to distinguish matching from non-matching image--text pairs and has little signal for the completeness and wording of captions that already match the image. The effect is not a weakness of a particular encoder: once references are added, RefCLIP-S and RefPAC-S recover most or all of the difference (\ref{app:full}).

\subsection{Two more quantities, each with a human reference}

We therefore report $\tau$ over all samples, for continuity with published tables, and $\tau$ above the lowest score, written $\tau(A)$. Where a set records several ratings per pair, we add a third quantity that asks the most practical question directly: given two candidates for the same image, does the metric prefer the one that people prefer?

\emph{Order accuracy.} For each pair with at least two ratings, we split its ratings at random into two halves of equal size, dropping one rating at random if their number is odd. Let $\bar y^{(1)}$ and $\bar y^{(2)}$ be the means of the two halves. For two candidates $a$ and $b$ of one image with $\bar y^{(1)}_a\neq\bar y^{(1)}_b$, a competitor $f$ scores
\begin{equation}
\mathrm{acc}(f;a,b)=\begin{cases}
1 & \text{if } \bigl(f_a-f_b\bigr)\bigl(\bar y^{(1)}_a-\bar y^{(1)}_b\bigr)>0,\\
\tfrac12 & \text{if } f_a=f_b,\\
0 & \text{otherwise,}
\end{cases}
\label{eq:order}
\end{equation}
and order accuracy is the mean over all such candidate pairs, over both assignments of the halves and over 20 random splits. Candidate pairs in which the first half ties are left out; a competitor that ties is credited with one half, the value of guessing.

\begin{figure}[!t]
\centering
\includegraphics[width=\linewidth]{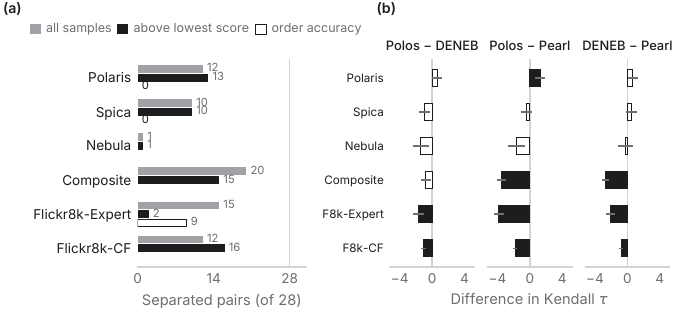}
\caption{Resolving power of the evaluation sets. (a)~Of the 28 pairs that can be formed from the eight best-ranked metrics of a set, the number that a paired bootstrap test separates after Holm correction, for each reported quantity. Order accuracy exists only where several ratings per pair are recorded; a 0 without a bar means that it separates none of the 28 pairs, as on the Polaris and Spica test sets. (b)~Paired differences between the three learned regression metrics in $\tau$ over all samples (whiskers: 95\% interval); filled bars are significant after correction.}
\label{fig:power}
\end{figure}

\emph{Human reference.} The agreement between annotators is a natural point of reference for the agreement of a metric with people. In Eq.~\eqref{eq:order}, the human reference is the second half itself, $f=\bar y^{(2)}$, which holds exactly as much information as the first. For $\tau(A)$, the human reference replaces the metric with a held-out rating: for a sample $i$ with rating $y_i$ whose pair has further ratings,
\begin{equation}
h_i=\frac{1}{|R_i|-1}\sum_{j\in R_i,\,j\neq i}y_j,\qquad \tau_{\text{human}}=\tau_c\bigl(h,\,y\bigr),
\label{eq:human}
\end{equation}
where $R_i$ is the set of ratings for the pair of sample $i$. Metrics are compared with $\tau_{\text{human}}$ on the same samples by a paired test.

\subsection{What the test sets can resolve}

A single interval per metric does not say which differences a benchmark supports. We score all metrics on the same bootstrap draws and test every pair (Section~\ref{sec:tools}). Fig.~\ref{fig:power}a counts the separated pairs among the eight best metrics of each set, where the interesting comparisons are. The Polaris and Spica test sets separate fewer than half of them, and the Nebula test set, the smallest, almost none. Order accuracy on Polaris and Spica separates no pair at all, because these sets contain only a few hundred pairs of candidates for the same image with different human scores, against more than fourteen thousand in Flickr8k-Expert.

Fig.~\ref{fig:power}b shows the consequence for the three learned regression metrics. On the test sets of their own family, only \LearnedClearFamily{} of nine pairwise differences is significant, and the sign changes from set to set. On the earlier benchmarks \LearnedClearEarlier{} of nine are significant and the order is always Pearl, DENEB, Polos. The recent test sets are large, but much of their size lies in lowest-score samples and in repeated ratings of the same pairs, which add little power to rank metrics.

\subsection{Metrics and the human reference}

Above the lowest score, most metrics are significantly below the human reference of Eq.~\eqref{eq:human}: \BelowPolaris{} of \JudgedPolaris{} on Polaris and \BelowSpica{} of \JudgedSpica{} on Spica. Three are significantly above it on both sets, the zero-shot judges that see the references (\ref{app:full}). None of them was trained on these scores. The human reference is therefore a point of orientation and not a ceiling: a single rating is itself a noisy estimate of the mean opinion, and a metric that tracks the mean can agree with a rating better than another single rating does. On Flickr8k-Expert, no metric is significantly above the reference.

\FloatBarrier
\section{A protocol without duplicates, and a re-ranking}
\label{sec:protocol}

\subsection{Protocol}

\begin{table}[!t]
\caption{The evaluation protocol.}
\label{tab:protocol}
\centering\tablefont
\begin{tabular}{@{}p{0.6cm}p{10.5cm}@{}}
\toprule
\textbf{P1} & \textbf{Rank only without duplicates.} A learned metric is ranked on an evaluation set with a model whose training set holds no sample of that set: the released model if its training set qualifies, otherwise a model retrained on the deduplicated training set. If neither exists, the metric is listed but not ranked. \\[3pt]
\textbf{P2} & \textbf{Report detection and ranking separately.} Give $\tau$ over all samples and above the lowest score, and order accuracy where several ratings per pair are recorded. \\[3pt]
\textbf{P3} & \textbf{Test pairs of metrics.} Score all metrics on the same bootstrap draws of images, correct for multiple comparisons, and state which metrics are not separated from the first. \\[3pt]
\textbf{P4} & \textbf{Average seeds and state the checkpoint rule.} Train a learned metric at least three times and say how the checkpoint was chosen. \\[3pt]
\textbf{P5} & \textbf{Show the human reference on the same samples}, as orientation and not as an upper bound. \\
\bottomrule
\end{tabular}
\end{table}

Table~\ref{tab:protocol} turns the findings into five rules. P1 follows from Section~\ref{sec:inflation}: one cannot know in advance which model memorizes, so duplicates are excluded by construction; deduplicated models are used only where needed because deduplication has a small cost elsewhere. P2 follows from Eq.~\eqref{eq:components} and P3 from Fig.~\ref{fig:power}. P4 reflects how unstable one training can be: on Flickr8k-Expert the epochs of a single Pearl training differ by up to \ExpertSpreadHigh{} points. For P1, we release, for every training set and evaluation set, the list of duplicated samples.

\subsection{Re-ranking}

We apply the protocol to \CompositeRanked{} metrics: six text-overlap metrics and a text-only baseline~\cite{gao2021simcse}; CLIP-S, PAC-S and PAC-S++ with two backbones, with and without references; Polos, DENEB and Pearl, retrained with three seeds where a deduplicated training set is needed; a ridge regression on frozen SigLIP or CLIP features as a minimal learned baseline; Rigel; and three zero-shot judges built on the open models Qwen3-VL (4B and 8B)~\cite{bai2025qwen3vl} and InternVL3.5-8B~\cite{wang2025internvl35}. A judge receives the image, the candidate, and, in one of its two modes, the references, and is asked for a rating from 0 to 9 with the rubric of FLEUR~\cite{lee2024fleur}. Its score is the expected rating under the probabilities $p_d$ that the model assigns to the ten digits as the next token,
\begin{equation}
s=\frac{1}{9}\,\frac{\sum_{d=0}^{9}d\,p_d}{\sum_{d=0}^{9}p_d},
\label{eq:judge}
\end{equation}
which avoids the ties of a single printed digit (prompt and settings in \ref{app:details}).

\begin{figure}[!t]
\centering
\includegraphics[width=\linewidth]{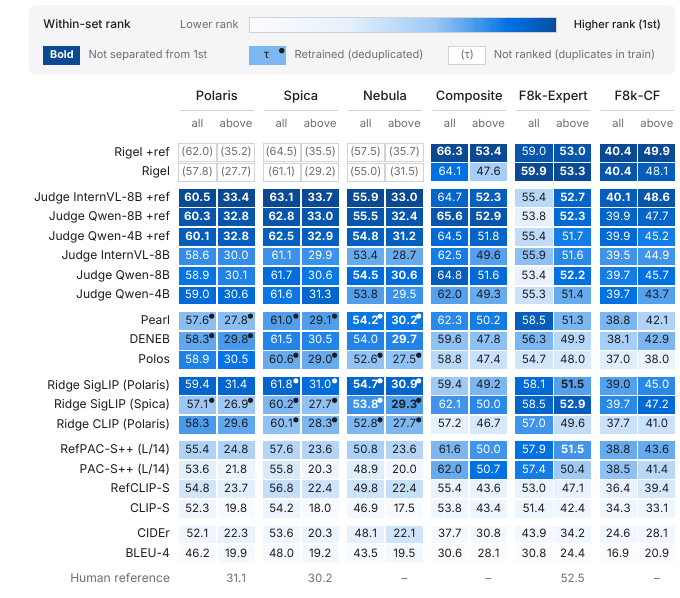}
\caption{Re-ranking under the protocol. Each cell gives $100\times$ Kendall's $\tau$ over all samples (``all'') or above the lowest human score (``above''); a darker cell is a better rank among the metrics ranked on that set. Bold: not separated from the first. A dot marks cells in which the ranked model was retrained on the deduplicated training set, because the released one holds samples of that evaluation set; cells without a dot use the released model. Values in parentheses on white are not ranked: only a model trained with samples of that set exists. Learned metrics are means of three seeds. The last row is the human reference of Eq.~\eqref{eq:human}. All \CompositeRanked{} metrics with rank intervals: \ref{app:full}.}
\label{fig:heatmap}
\end{figure}

\begin{figure}[!t]
\centering
\includegraphics[width=\linewidth]{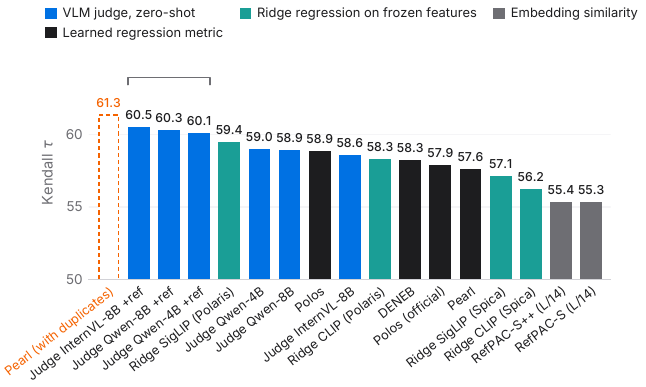}
\caption{The sixteen best-ranked metrics on the Polaris test set ($\tau$ over all samples; the vertical axis starts at 50), coloured by family. The bracket spans the metrics that are not separated from the first. The outlined bar is Pearl trained on the released training set, which holds \DupSpicaPolaris\% of the Polaris test samples; it is shown for comparison and not ranked.}
\label{fig:leaderboard}
\end{figure}

Fig.~\ref{fig:heatmap} gives the results for twenty representative metrics, and Fig.~\ref{fig:leaderboard} shows the leading group on the Polaris test set. We read five things from them.

\emph{Zero-shot judges lead on the recent test sets.} On Polaris and Spica, the metrics not separated from the first are exactly the three judges with references, over all samples and above the lowest score.

\emph{A ridge regression matches or exceeds the learned regression metrics.} Polos, DENEB and Pearl rank between \LearnedRankBest th and \LearnedRankWorst th on the recent test sets. A ridge regression on frozen SigLIP features, which is a single linear map with one tuned penalty, ranks \RidgeRecentRank th on all three and is ahead of each of them by \RidgeGapLow{} to \RidgeGapHigh{} points (teal bar in Fig.~\ref{fig:leaderboard}); several of these differences are within the resolution of the test sets (Section~\ref{sec:validity}). Over the six evaluation sets and the three metrics, it is significantly above in \RidgeAbove{} of \RidgeComparisons{} paired comparisons and significantly below in \RidgeBelow{} (Pearl on Composite). The inputs are not equal: the ridge regression uses the feature layout of Polos with SigLIP as its image--text encoder (\ref{app:details}), and the three metrics use other encoders. To separate the encoder from the architecture, we fit the same ridge regression to the CLIP features that Polos reads, on the same Polaris training set. On the three recent test sets, this CLIP ridge regression is not separated from Polos (differences of \RidgeClipPolosLow{} to \RidgeClipPolosHigh{} points); on the earlier benchmarks it is significantly above Polos on Flickr8k-Expert and Flickr8k-CF and below it on Composite. The architecture of Polos thus adds no consistent gain over a linear readout of its own features, whereas the newer encoder adds \RidgeEncoderGapLow{} to \RidgeEncoderGapHigh{} points to the same readout on the recent test sets: on these benchmarks, much of what a learned metric scores comes from its encoder. The regression metrics keep the advantage of cost over a judge, and Pearl is third on Flickr8k-Expert.

\emph{Fine-tuning the judge helps where it can be ranked.} Rigel is first on Flickr8k-Expert, Composite and Flickr8k-CF.\footnote{Strictly, P1 would exclude models trained on the released Spica and Nebula sets from Composite, as each holds one of its 11{,}982 samples. Because one sample changes $\tau$ by less than 0.04 points and Rigel cannot be retrained, we rank them on Composite.} On the three recent test sets, it cannot be ranked under P1, because only a checkpoint trained on the released Spica training set exists; its values there are slightly above the best ranked metric, and Section~\ref{sec:inflation} found no sign that duplicates produce this lead. Fine-tuning also repairs a weakness of the zero-shot judges, several of which are in the leading group of Flickr8k-Expert above the lowest score but all of which are only in the middle over all samples: on a set with many lowest-score samples they detect less sharply than they rank.

\emph{Training-free embedding metrics hold up on the earlier benchmarks.} PAC-S++ without references is among the best metrics on Composite and Flickr8k-Expert that use neither human scores nor a generative model.

\emph{Duplicates can change the first place.} Pearl trained on the released training set would be above every ranked metric on the Polaris test set in \PearlDupFirst\% of bootstrap draws; under the protocol, it ranks \PearlPolarisRank th (Fig.~\ref{fig:leaderboard}).

\subsection{Case studies}

\begin{figure}[!t]
\centering
\includegraphics[width=\linewidth]{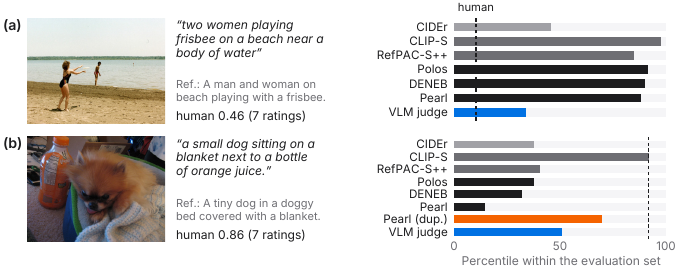}
\caption{Two pairs from the Polaris test set. Each row shows the image, the candidate (in quotation marks), one of the references, and the mean human score on a 0--1 scale. On the right, a bar gives the percentile at which a metric places the pair among all pairs of the set that are rated above the lowest score, and the dashed line gives the percentile of the human score. (a)~The candidate names two women where the image shows a woman and a man; learned and embedding metrics rate it high, the judge rates it low, as people do. (b)~A duplicated sample: Pearl trained without it underrates the pair, and Pearl trained with it (orange) moves most of the way to the human score. Twelve further cases: \ref{app:cases}.}
\label{fig:cases}
\end{figure}

Numbers over thousands of samples do not show what a metric gets right or wrong. Fig.~\ref{fig:cases} gives two pairs that represent patterns we found when reading the largest disagreements (\ref{app:cases} has twelve more, from three sets). In~(a), the candidate is fluent and close to the references in wording but wrong in one fact. Embedding and learned regression metrics place it near the top; the judge, which reads the image and the references together, places it low. This is the typical success of a judge. Judges are not uniformly better, however; they are better on average and fail differently, for example, by underrating a correct caption on which the other metrics agree with people. Case~(b) shows inflation on a single sample. Pearl trained on the deduplicated set places the pair far below the human score; Pearl trained on the released set, which holds the pair with its seven ratings, places it much closer. Caption and image are unchanged.

\subsection{Judges behind a proprietary interface}

We evaluate judges built on open models and do not run CLAIR, G-VEval, or other judges that call a proprietary model. There are three reasons, in order of weight. First, a protocol must be repeatable: hosted models are updated and withdrawn, so a number measured today cannot be reproduced or extended to a new evaluation set later. Second, Eq.~\eqref{eq:judge} needs the probabilities of all ten digits, which hosted interfaces expose only in part; Kim et al.~\cite{kim2025expert} report the same limitation for GPT-4o. Third, the training data of a proprietary model cannot be audited, which is the very question of this paper. Published results indicate what such judges would add: Kim et al. report for a reference-free GPT-4o judge 58.2 on the Polaris test set, 54.3 on the Nebula test set and 65.9 on Composite, close to the range of our open reference-free judges (Fig.~\ref{fig:heatmap}). These figures come from another implementation and were not measured under our protocol, so we do not rank them; a repeatable comparison with proprietary judges remains open.

\FloatBarrier
\section{Discussion}
\label{sec:discussion}

\paragraph{Beyond this family of benchmarks} Nothing in the mechanism is specific to captioning. A collection of human judgments is expensive, so a new collection reuses the images, systems, and sometimes the judgments of an earlier one; a learned evaluator is then tested on both. The same holds for learned metrics in machine translation, summarization and image quality assessment. Three elements of this study transfer directly. Matching by content, not by identifier, finds duplicates that survive renaming. The comparison within a test set, by groups and by epoch, separates memorization from the many other consequences of changing a training set. And the decomposition of a rank correlation into detection and ranking applies to any benchmark whose scale has a heavily populated lowest level. More broadly, the reliability of a benchmark is an empirical property that can be measured, and measuring it belongs to the evaluation of a pattern recognition method as much as the headline number does.

\paragraph{Caveats on VLM judges} The pretraining data of the judges is not documented at the level of individual images, so we cannot exclude that they have seen COCO or Flickr images with their reference captions; human scores of candidate captions are a far less likely input. All four judges, Rigel included, build on language models of one generation, and judges based on large language models can favour text produced by similar models~\cite{panickssery2024self}; the benchmarks studied here cannot test this. A judge is also orders of magnitude slower than a regression metric.

\paragraph{Limitations} The comparison of released and deduplicated training sets is not a randomized experiment, and deduplication has a general cost of about one point whose source we can only make plausible (Section~\ref{sec:checks}); our claims of inflation rest on the within-set, matched and per-epoch comparisons, which subtract it. For Rigel we can report only that no inflation was detected. Whether sharing an image without its candidate has a small effect remains open (Section~\ref{sec:imageonly}). We study one family of collections, and three of the six evaluation sets give no human reference.

\section{Conclusion}

The three recent collections of human judgments for image captioning made learned metrics possible. Because they were built on one another, samples from their test sets, with the same ratings, are present in the training sets of sibling collections. Whether this inflates a metric depends on the learner and on how often a training set repeats a sample: for Pearl, the gain on duplicated samples grows with every epoch; for DENEB, it is below one point. Independently of duplication, most of the correlation that benchmark tables report measures the detection of captions with the lowest human score, an ability that current metrics share, and the recent test sets do not order the learned regression metrics of their own family; a ridge regression on frozen features matches or exceeds all three there. With duplicates excluded by construction and the two abilities reported separately, zero-shot VLM judges rank first on the recent test sets and a fine-tuned judge on the earlier benchmarks. We hope that the protocol, the filter lists, and the statistical tools assembled here will be useful for the next generation of learned evaluators, in captioning and beyond.


\section*{CRediT authorship contribution statement}

\textbf{Xiaochen Zhang}: Conceptualization, Methodology, Software, Validation, Formal analysis, Investigation, Data curation, Visualization, Writing -- original draft. \textbf{Jin Li}: Conceptualization, Methodology, Validation, Formal analysis, Data curation, Writing – original draft, Writing – review \& editing. \textbf{Chunlong Xia}: Software, Writing -- original draft. \textbf{Longjiang Guo}: Conceptualization, Methodology, Resources, Supervision, Project administration, Funding acquisition, Validation, Writing – review \& editing.

\section*{Declaration of competing interest}
The authors declare that they have no known competing financial interests or personal relationships that could have appeared to influence the work reported in this paper.

\section*{Funding}
This work was supported in part by the National Natural Science Foundation of China under Grant No. 61977044, and by the University-Industry Collaborative Education Program of the Ministry of Education of the People's Republic of China under Grant Nos. 250405448235212 and 250405448255034. The funders had no role in study design, data collection and analysis, interpretation of the results, writing of the manuscript, or the decision to submit it for publication. The funders had no role in study design, data collection and analysis, interpretation of the results, writing of the manuscript, or the decision to submit it for publication.

\section*{Data availability}
The filter lists (pixel hashes of images and indices of duplicated samples for every training set and evaluation set, under the exact and the relaxed matching rule) and all analysis scripts are available at \url{https://github.com/morninghut/ic-dataset-audit}. The collections themselves are available from their authors; we do not redistribute images, captions or annotations.

\section*{Declaration of generative AI and AI-assisted technologies in the manuscript preparation process}
During the preparation of this work the authors used Codex and AmpCode in order to assist in writing code, including the plotting code of the figures, and to polish the language of the manuscript. After using this tool/service, the authors reviewed and edited the content as needed and take full responsibility for the content of the published article.

\appendix
\renewcommand{\appendixname}{Appendix}
\counterwithin*{table}{section}
\counterwithin*{figure}{section}
\ifdefined\WithoutAppendix\else
\newcommand{\apptable}[3]{%
  \begin{table}[!htbp]\caption{#2}\label{#1}\centering\apptablefont\makebox[\linewidth]{\input{generated/#3}}\end{table}}

\section{Reproduction of published results}
\label{app:repro}

Table~\ref{tab:repro} compares our measurements with the numbers printed in the papers of the metrics, on the sets where both exist and the training set is the released one. Polos and Rigel are the released checkpoints; DENEB and Pearl are our retrainings from the public code with the public configuration (mean of three seeds). Agreement is within half a point except on Composite, where the retrained DENEB and Pearl are 1.4 and 1.9 points \emph{above} the published values; we do not know why. The text-overlap metrics agree with the table of the Polos paper to within 0.1 points, with one exception (BLEU-4 on Flickr8k-CF, 0.5).

  \begin{table}[!htbp]\caption{Measured / published $100\times\tau$. All models are trained on the released training set, also where the main ranking uses a deduplicated model or none.}\label{tab:repro}\centering\apptablefont\makebox[\linewidth]{\begin{tabular}{lrrrrr}
\toprule
Metric & Flickr8k-Expert & Flickr8k-CF & Composite & Nebula test & Polaris test \\
\midrule
DENEB & 56.3 / 56.8 & 38.1 / 38.3 & 59.6 / 58.2 & 54.0 / 54.3 & -- \\
Pearl & 58.5 / 58.6 & 38.8 / 38.6 & 62.3 / 60.4 & 54.9 / 55.4 & -- \\
Rigel & 59.9 / 59.9 & 40.4 / 40.4 & 64.1 / 64.3 & 55.0 / 55.1 & -- \\
Rigel +\,ref & 59.0 / 59.0 & 40.4 / 40.4 & 66.3 / 66.1 & 57.5 / 57.5 & -- \\
Polos (official) & 56.4 / 56.4 & 37.8 / 37.8 & 58.1 / 57.6 & -- & 57.9 / 57.8 \\
\bottomrule
\end{tabular}
}\end{table}

Three implementation details differ from the public pipelines; none changes a conclusion. (i)~Pearl's training code looks up one text feature with the raw candidate and its evaluation code with a prompt prefix; scoring both ways changes no number by more than 0.25 points. (ii)~The public feature extraction of Pearl pads a batch of texts for one of its encoders without a mask, so a feature depends on the other texts of its batch; we extract every text on its own, which is reproducible. (iii)~DENEB's batching truncates the references of a batch to the smallest count in it; we group samples by reference count so that each keeps all its references.

\section{Duplication counts, label agreement and near duplicates}
\label{app:matrix}

Table~\ref{tab:matrix} gives the counts behind Fig.~\ref{fig:duplication} for every training set and every split we examined, and Table~\ref{tab:sets} lists the evaluation sets with the quantities used in Sections~\ref{sec:validity} and~\ref{sec:protocol}. The upper bound in Table~\ref{tab:matrix} adds samples whose image matches a training image with a cosine similarity above 0.97 in SigLIP feature space and whose candidate equals a training candidate of that image after lower-casing and removing punctuation, or has a character similarity above 0.9. At the stricter image threshold of 0.99, which by inspection corresponds to re-encoded copies of one photograph (Fig.~\ref{fig:examples}b), the Spica training set contains copies of \SpicaReencoded{} Spica test images, which touch \SpicaReencodedRows{} test samples.

  \begin{table}[!htbp]\caption{Samples of each split that are duplicated in each training set: exact matches, their share, the upper bound with near duplicates, and the share of image-only samples.}\label{tab:matrix}\centering\apptablefont\makebox[\linewidth]{\begin{tabular}{lrrrrr}
\toprule
& & \multicolumn{3}{c}{Duplicated samples} & Image \\
\cmidrule(lr){3-5}
Training set / evaluation set & Samples & exact & \% & \% upper & only \% \\
\midrule
\multicolumn{6}{l}{\emph{Polaris train}} \\
\quad Polaris test & 26,122 & 0 & 0.0 & 0.0 & 0.0 \\
\quad Polaris val & 26,268 & 0 & 0.0 & 0.0 & 0.0 \\
\quad Spica test & 15,238 & 6,603 & 43.3 & 43.3 & 11.7 \\
\quad Spica val & 8,773 & 2,374 & 27.1 & 27.3 & 12.7 \\
\quad Nebula test & 3,298 & 806 & 24.4 & 24.5 & 1.3 \\
\quad Nebula val & 3,298 & 740 & 22.4 & 22.4 & 1.5 \\
\quad Flickr8k-Expert & 16,992 & 0 & 0.0 & 0.0 & 0.0 \\
\quad Composite & 11,982 & 0 & 0.0 & 0.5 & 16.0 \\
\quad Flickr8k-CF & 47,830 & 0 & 0.0 & 0.0 & 0.0 \\
\quad PASCAL-50S & 40,000 & 0 & 0.0 & 0.0 & 0.0 \\
\quad FOIL & 198,960 & 8 & $<$0.1 & 0.2 & 17.2 \\
\addlinespace
\multicolumn{6}{l}{\emph{Nebula train}} \\
\quad Polaris test & 26,122 & 13,111 & 50.2 & 51.1 & 29.6 \\
\quad Polaris val & 26,268 & 13,130 & 50.0 & 50.6 & 29.6 \\
\quad Spica test & 15,238 & 0 & 0.0 & $<$0.1 & 0.0 \\
\quad Spica val & 8,773 & 3,664 & 41.8 & 42.5 & 33.4 \\
\quad Nebula test & 3,298 & 0 & 0.0 & 0.1 & 0.0 \\
\quad Nebula val & 3,298 & 0 & 0.0 & $<$0.1 & 0.0 \\
\quad Flickr8k-Expert & 16,992 & 0 & 0.0 & 0.0 & 0.0 \\
\quad Composite & 11,982 & 1 & $<$0.1 & 0.7 & 33.5 \\
\quad Flickr8k-CF & 47,830 & 0 & 0.0 & 0.0 & 0.0 \\
\quad PASCAL-50S & 40,000 & 0 & 0.0 & 0.0 & 0.0 \\
\quad FOIL & 198,960 & 13 & $<$0.1 & 0.4 & 57.9 \\
\addlinespace
\multicolumn{6}{l}{\emph{Spica train}} \\
\quad Polaris test & 26,122 & 17,856 & 68.4 & 68.6 & 22.9 \\
\quad Polaris val & 26,268 & 18,108 & 68.9 & 69.1 & 23.3 \\
\quad Spica test & 15,238 & 342 & 2.2 & 3.8 & 38.0 \\
\quad Spica val & 8,773 & 190 & 2.2 & 3.3 & 39.7 \\
\quad Nebula test & 3,298 & 4 & 0.1 & 1.3 & 39.1 \\
\quad Nebula val & 3,298 & 3,148 & 95.5 & 95.5 & 1.9 \\
\quad Flickr8k-Expert & 16,992 & 0 & 0.0 & 0.0 & 0.0 \\
\quad Composite & 11,982 & 1 & $<$0.1 & 1.7 & 47.6 \\
\quad Flickr8k-CF & 47,830 & 0 & 0.0 & 0.0 & 0.0 \\
\quad PASCAL-50S & 40,000 & 0 & 0.0 & 0.0 & 0.0 \\
\quad FOIL & 198,960 & 20 & $<$0.1 & 0.6 & 85.0 \\
\bottomrule
\end{tabular}
}\end{table}

  \begin{table}[!htbp]\caption{Evaluation sets: samples, share with the lowest score, human reference $\tau$ above the lowest score (Eq.~\eqref{eq:human}), number of candidate pairs of one image with different human scores (the items of order accuracy), and share of duplicated samples in each training set.}\label{tab:sets}\centering\apptablefont\makebox[\linewidth]{\begin{tabular}{lrrrrrrr}
\toprule
& & Lowest & Human & Same-image & \multicolumn{3}{c}{Duplicated samples (\%) in} \\
\cmidrule(l){6-8}
Evaluation set & Samples & score (\%) & $\tau$ & pairs & Polaris tr. & Nebula tr. & Spica tr. \\
\midrule
Polaris test & 26,122 & 23.7 & 31.1 & 539 & 0 & 50.2 & 68.4 \\
Spica test & 15,238 & 26.3 & 30.2 & 308 & 43.3 & 0 & 2.2 \\
Nebula test & 3,298 & 26.2 & -- & -- & 24.4 & 0 & 0.1 \\
\addlinespace
Flickr8k-Expert & 16,992 & 57.7 & 52.5 & 14,605 & 0 & 0 & 0 \\
Composite & 11,982 & 28.3 & -- & -- & 0 & $<$0.1 & $<$0.1 \\
Flickr8k-CF & 47,830 & 87.7 & -- & -- & 0 & 0 & 0 \\
\bottomrule
\end{tabular}
}\end{table}

Table~\ref{tab:labels} compares the two copies of each duplicated pair. Because one side may store every rating and the other their mean, both sides are averaged per pair first. The copies agree exactly for nearly all pairs. Compared without this averaging, a single rating of the Polaris test set equals the mean stored in the Nebula training set for only a few percent of the samples, which is a property of the storage format and not a difference of opinion.

  \begin{table}[!htbp]\caption{The duplicated part of each test set counted in samples and in pairs, the number of training samples per duplicated pair, the share of duplicated pairs whose mean rating is identical in both sets, and the correlation $r$ of the two means.}\label{tab:labels}\centering\apptablefont\makebox[\linewidth]{\begin{tabular}{lrrrrrrr}
\toprule
& \multicolumn{2}{c}{Duplicated samples} & \multicolumn{2}{c}{Duplicated pairs} & Training & Same & \\
\cmidrule(lr){2-3} \cmidrule(lr){4-5}
Training $\rightarrow$ test set & count & \% & count & \% & samples per pair & mean (\%) & $r$ \\
\midrule
Spica $\rightarrow$ Polaris & 17,856 & 68.4 & 2,787 & 31.9 & 6.4 & 96.7 & 0.990 \\
Nebula $\rightarrow$ Polaris & 13,111 & 50.2 & 2,057 & 23.6 & 1.0 & 99.9 & 1.000 \\
Polaris $\rightarrow$ Spica & 6,603 & 43.3 & 1,030 & 16.0 & 6.4 & 100.0 & 1.000 \\
Polaris $\rightarrow$ Nebula & 806 & 24.4 & 806 & 24.4 & 6.4 & 100.0 & 1.000 \\
\bottomrule
\end{tabular}
}\end{table}

\section{The released Spica collection}
\label{app:spica}

Four properties of the public release matter for anyone who trains or tests on it.
\begin{itemize}
\item \emph{One image, two identifiers.} The \SpicaIds{} image identifiers refer to \SpicaPictures{} distinct images; the others are byte-identical files under a second identifier. The splits are disjoint by identifier but not by content: \SpicaTestInTrain{} of the \SpicaTestPictures{} test images (\SpicaTestInTrainShare\%) are also training images.
\item \emph{The test set contains Nebula's.} \SpicaNebulaCommonPairs{} of the Nebula test pairs are in the Spica test set with their individual ratings (\SpicaCommonRatings{} samples), and averaging these ratings reproduces the Nebula score in 99.9\% of the samples. 
The Spica test set holds \SpicaOnlyPairs{} further pairs, \SpicaOnlyLowest\% of which carry a single rating with the lowest score.
\item \emph{Half of the training samples have no reference.} \SpicaNoRefs{} of the \SpicaRows{} released samples carry an empty reference list, all of them in the training split (48\% of it).
\item \emph{Own-split duplicates.} \DupRowsSpicaSpica{} Spica test samples (\SpicaOwnDup\%; at most \SpicaOwnDupUpper\% with near duplicates) are duplicated in the Spica training split.
\end{itemize}

\section{Implementation details}
\label{app:details}

\paragraph{Image and caption matching} An image file is decoded with Pillow and converted to 8-bit RGB; no rotation from EXIF metadata is applied. Its hash is the SHA-1 digest of the string ``\emph{width}x\emph{height}|'' followed by the raw RGB bytes. Two files with the same hash therefore show identical pixels, whatever their names, formats or metadata. Candidates are compared as exact strings. The released filter lists contain, for each training set and evaluation set, the hashes of the images and the indices of the duplicated samples; they contain no image, caption or score.

\paragraph{Bootstrap and tests} Intervals are percentile intervals over 10{,}000 draws of images with replacement (1{,}000 for the group analyses of Section~\ref{sec:inflation} and the cross-fitting). For a learned metric the draws of its three seeds are averaged before the interval is taken. The $p$-value of a paired difference is computed from a normal approximation with the bootstrap standard error, because a percentile $p$-value cannot fall below two over the number of draws, and is then corrected by Holm's procedure over all pairs of metrics of an evaluation set. Order accuracy uses 20 random splits of the ratings and draws images in the same way.

\paragraph{Ridge regression and network} Both use frozen features in the layout of Polos: from a sentence encoder and from a vision--language encoder, the embeddings of the candidate and of a reference, their element-wise product and their absolute difference; and from the vision--language encoder also the product and the absolute difference of the candidate and image embeddings. These ten groups are concatenated. The sentence encoder is \texttt{sup-simcse-roberta-base}~\cite{gao2021simcse} (a uniform mixture of its 13 layers); the vision--language encoder is SigLIP \texttt{so400m-patch14-384}, which gives 9{,}984 dimensions, or CLIP ViT-B/32 for the rows marked CLIP, which gives 6{,}144. One feature vector is built per reference; in training each reference is a weighted example so that every sample counts once, and at test time the score of a sample is the maximum over its references. The ridge penalty is chosen from $\{0.1, 0.3, 1, 3, 10, 30\}$ by $\tau$ on validation images that are in neither the training set nor any test set. The network has layers of 9{,}984, 2{,}304, 1{,}152 and 1 units with tanh activations and dropout 0.1, and is trained with a learning rate of $3\times10^{-5}$, batches of 64 with four accumulation steps, and early stopping with a patience of five epochs. ``Polos architecture'' is the public Polos model trained on cached CLIP features. Folds for cross-fitting are assigned by image.

\paragraph{Zero-shot judges} \sloppy The judges are \texttt{Qwen/Qwen3-VL-4B-Instruct} (revision \texttt{ebb281ec}), \texttt{Qwen/Qwen3-VL-8B-Instruct} (\texttt{0c351dd0}) and \texttt{OpenGVLab/InternVL3\_5-8B-HF} (\texttt{741a7d03}); Rigel is the released checkpoint (\texttt{303d2086}). All run with \texttt{transformers} 4.57.1 and \texttt{torch} 2.5.1 in bfloat16. An image is passed at the resolution that yields at most 256 visual tokens. In the mode with references, the first five references in the stored order are used. The prompt is the following; the bracketed part is included only in the mode with references.

\begin{quote}\footnotesize\linespread{1}\selectfont
Your task is to evaluate and rate the caption on a scale of 0 to 9 based on the given Grading Criteria. (Print the score ONLY)\\
Grading Criteria:\\
0: The caption does not describe the image at all.\\
9: The caption accurately and clearly describes the image.\\
{[}Reference captions written by people for this image: \ldots{]}\\
Caption: \ldots\\
Score (Choose a rating from 0 to 9):
\end{quote}

The score is Eq.~\eqref{eq:judge}, computed from the probabilities of the ten digit tokens at the first generated position; each digit is a single token in all three tokenizers. For the Qwen3-VL judges these tokens carry nearly all of the probability mass. InternVL3.5 sometimes opens a reasoning block first, so we prefill the empty block ``\texttt{<think>\textbackslash n\textbackslash n</think>\textbackslash n\textbackslash n}'' and read the rating after it. Reading the score without the prefill, from the renormalized digit probabilities, changes $\tau$ by at most \NothinkRestMax{} points on five sets but by \NothinkExpertLow{} to \NothinkExpertHigh{} points on Flickr8k-Expert (Table~\ref{tab:nothink}). Details of inference can thus move a judge by more than the differences between judges, and they need to be reported.

  \begin{table}[!htbp]\caption{InternVL3.5-8B: $100\times\tau$ over all samples with the score read from renormalized digit probabilities (conditional) and after prefilling an empty reasoning block (prefilled, used in the paper).}\label{tab:nothink}\centering\apptablefont\makebox[\linewidth]{\begin{tabular}{lrrrrrr}
\toprule
& \multicolumn{3}{c}{With references} & \multicolumn{3}{c}{Without references} \\
\cmidrule(lr){2-4} \cmidrule(l){5-7}
Evaluation set & conditional & prefilled & diff. & conditional & prefilled & diff. \\
\midrule
Polaris test & 60.51 & 60.49 & $-$0.02 & 58.83 & 58.57 & $-$0.26 \\
Spica test & 62.92 & 63.11 & $+$0.20 & 61.39 & 61.11 & $-$0.28 \\
Nebula test & 55.56 & 55.87 & $+$0.31 & 53.57 & 53.39 & $-$0.17 \\
Flickr8k-Expert & 52.99 & 55.43 & $+$2.44 & 52.92 & 55.88 & $+$2.96 \\
Composite & 64.13 & 64.67 & $+$0.55 & 62.73 & 62.48 & $-$0.24 \\
Flickr8k-CF & 39.83 & 40.14 & $+$0.31 & 39.24 & 39.53 & $+$0.30 \\
\bottomrule
\end{tabular}
}\end{table}

\section{Cross-fitting in full}
\label{app:crossfit}

Table~\ref{tab:crossfitfull} extends Table~\ref{tab:crossfit} to all groups of samples, over all samples and above the lowest score. For samples whose image is not in the training set, all five models are in the same position, so the ``unseen image'' columns are a placebo and measure the noise between models. The image-only columns compare models that saw the image with other candidates against a model that did not see the image; over all samples they are zero for the reason given in Section~\ref{sec:imageonly}, and above the lowest score their intervals include zero. The released Polos checkpoint cannot be cross-fitted; a difference-in-differences against our unseen models gives no significant inflation for it on the Nebula and Spica test sets. 
\MlpNebulaNote

  \begin{table}[!htbp]\caption{Cross-fit effect $E_g$ (Eq.~\eqref{eq:crossfit}) with 95\% interval, by group of samples, over all samples and above the lowest score. Dashes: fewer than 300 samples. An asterisk marks a network trained with one seed.}\label{tab:crossfitfull}\centering\apptablefont\makebox[\linewidth]{\begin{tabular}{lrrrrrr}
\toprule
& \multicolumn{2}{c}{Duplicated} & \multicolumn{2}{c}{Image only} & \multicolumn{2}{c}{Unseen image} \\
\cmidrule(lr){2-3} \cmidrule(lr){4-5} \cmidrule(l){6-7}
Learner, training $\rightarrow$ test set & all & above & all & above & all & above \\
\midrule
\multicolumn{7}{@{}l}{\emph{Ridge regression, SigLIP features}} \\
\quad Spica $\rightarrow$ Polaris & $+$2.1 [$+$2.0, $+$2.4] & $+$2.2 [$+$2.0, $+$2.4] & $+$0.0 [$-$0.0, $+$0.0] & $+$0.1 [$-$0.6, $+$0.7] & $-$0.4 [$-$1.0, $-$0.1] & $-$0.9 [$-$1.8, $-$0.1] \\
\quad Nebula $\rightarrow$ Polaris & $+$2.0 [$+$1.7, $+$2.3] & $+$2.1 [$+$1.8, $+$2.4] & $+$0.1 [$-$0.0, $+$0.2] & $+$0.2 [$-$0.4, $+$0.7] & $+$0.0 [$-$0.3, $+$0.2] & $-$0.0 [$-$0.5, $+$0.4] \\
\quad Polaris $\rightarrow$ Spica & $+$3.2 [$+$2.7, $+$3.7] & $+$3.3 [$+$2.8, $+$3.8] & $+$0.0 [$+$0.0, $+$0.0] & -- & $-$0.0 [$-$0.2, $+$0.1] & $-$0.1 [$-$0.5, $+$0.2] \\
\quad Polaris $\rightarrow$ Nebula & $+$5.6 [$+$4.9, $+$6.3] & $+$5.6 [$+$4.9, $+$6.4] & -- & -- & $-$0.1 [$-$0.2, $+$0.1] & $-$0.1 [$-$0.5, $+$0.2] \\
\multicolumn{7}{@{}l}{\emph{Ridge regression, CLIP features}} \\
\quad Spica $\rightarrow$ Polaris & $+$2.5 [$+$2.2, $+$2.7] & $+$2.5 [$+$2.2, $+$2.8] & $-$0.0 [$-$0.1, $+$0.0] & $+$0.2 [$-$0.8, $+$1.3] & $-$0.0 [$-$0.3, $+$0.3] & $-$0.0 [$-$0.6, $+$0.6] \\
\quad Polaris $\rightarrow$ Spica & $+$4.1 [$+$3.5, $+$4.9] & $+$4.2 [$+$3.5, $+$4.9] & $+$0.0 [$+$0.0, $+$0.0] & -- & $-$0.1 [$-$0.3, $+$0.2] & $-$0.2 [$-$0.6, $+$0.3] \\
\quad Polaris $\rightarrow$ Nebula & $+$6.7 [$+$5.7, $+$7.6] & $+$6.7 [$+$5.8, $+$7.6] & -- & -- & $-$0.0 [$-$0.2, $+$0.2] & $-$0.1 [$-$0.5, $+$0.3] \\
\multicolumn{7}{@{}l}{\emph{Three-layer network, SigLIP features}} \\
\quad Spica $\rightarrow$ Polaris & $+$8.4 [$+$7.7, $+$9.1] & $+$8.7 [$+$8.0, $+$9.5] & $+$0.0 [$-$0.1, $+$0.1] & $-$0.2 [$-$2.0, $+$2.0] & $-$0.2 [$-$1.1, $+$0.8] & $-$0.5 [$-$2.2, $+$1.4] \\
\quad Nebula $\rightarrow$ Polaris & $+$3.2 [$+$2.5, $+$3.8] & $+$3.3 [$+$2.6, $+$4.0] & $+$0.1 [$-$0.1, $+$0.3] & $+$0.5 [$-$0.7, $+$1.8] & $-$0.3 [$-$0.9, $+$0.3] & $-$0.7 [$-$1.8, $+$0.4] \\
\quad Polaris $\rightarrow$ Spica & $+$10.7 [$+$9.4, $+$12.1] & $+$10.9 [$+$9.6, $+$12.2] & $-$0.0 [$-$0.0, $-$0.0] & -- & $+$0.2 [$-$0.3, $+$0.6] & $+$0.1 [$-$0.7, $+$1.1] \\
\quad Polaris $\rightarrow$ Nebula & $+$21.4 [$+$19.4, $+$23.4] & $+$21.4 [$+$19.5, $+$23.6] & -- & -- & $+$0.1 [$-$0.4, $+$0.5] & $+$0.2 [$-$0.8, $+$1.2] \\
\multicolumn{7}{@{}l}{\emph{Polos architecture, CLIP features}} \\
\quad Polaris $\rightarrow$ Spica & $+$6.3 [$+$5.2, $+$7.3] & $+$6.4 [$+$5.3, $+$7.5] & $+$0.0 [$-$0.0, $+$0.0] & -- & $-$0.2 [$-$0.6, $+$0.3] & $-$0.4 [$-$1.3, $+$0.5] \\
\quad Polaris $\rightarrow$ Nebula & $+$10.9 [$+$8.9, $+$12.6] & $+$10.9 [$+$9.0, $+$12.7] & -- & -- & $-$0.1 [$-$0.6, $+$0.5] & $-$0.5 [$-$1.6, $+$0.6] \\
\bottomrule
\end{tabular}
}\end{table}

\section{Retraining contrasts and per-epoch scores}
\label{app:epochs}

Table~\ref{tab:retrain} lists all contrasts between training sets and Table~\ref{tab:groupsfull} repeats them above the lowest score, with the composition of each group. Tables~\ref{tab:epochs} and~\ref{tab:denebepochs} give the per-epoch means behind Fig.~\ref{fig:epochs}. The matched removal deletes whole pairs that are duplicated in no evaluation set, stratified by presence of references, number of ratings and mean score, until it has removed exactly as many samples as the deduplication. Training runs stop by the early-stopping rule of the public code, so they contribute different numbers of epochs.

  \begin{table}[!htbp]\caption{Differences between training sets by group of samples (mean of three seeds; 95\% bootstrap interval over images).}\label{tab:retrain}\centering\apptablefont\makebox[\linewidth]{\begin{tabular}{lrrrr}
\toprule
Samples & $n$ & \multicolumn{3}{c}{Difference in $\tau$ [95\% CI], mean of three seeds} \\
\midrule
\multicolumn{2}{l}{\emph{Pearl, Spica training set}} & released $-$ dedup. & matched $-$ dedup. & released $-$ matched \\
Polaris test, all samples & 26,122 & $+$3.72 [$+$3.39, $+$4.04] & $+$4.13 [$+$3.76, $+$4.49] & $-$0.42 [$-$0.60, $-$0.21] \\
\quad duplicated & 17,856 & $+$6.45 [$+$5.88, $+$7.04] & $+$7.37 [$+$6.70, $+$8.04] & $-$0.92 [$-$1.22, $-$0.62] \\
\quad image only & 5,987 & $+$0.07 [$+$0.03, $+$0.13] & $+$0.08 [$+$0.02, $+$0.16] & $-$0.01 [$-$0.07, $+$0.06] \\
\quad unseen image & 2,279 & $+$1.20 [$+$0.16, $+$2.26] & $+$0.18 [$-$0.78, $+$1.11] & $+$1.02 [$+$0.14, $+$1.95] \\
\quad duplicated $-$ unseen & & $+$5.24 [$+$4.08, $+$6.42] & $+$7.19 [$+$6.09, $+$8.31] & $-$1.94 [$-$2.92, $-$0.98] \\
Spica test, all samples & 15,238 & $+$1.08 [$+$0.77, $+$1.38] & $+$0.36 [$+$0.06, $+$0.66] & $+$0.72 [$+$0.45, $+$0.99] \\
Nebula test, all samples & 3,298 & $+$0.66 [$+$0.19, $+$1.11] & $+$0.48 [$+$0.03, $+$0.94] & $+$0.18 [$-$0.15, $+$0.53] \\
Composite, all samples & 11,982 & $+$1.11 [$+$0.93, $+$1.29] & $+$1.20 [$+$0.99, $+$1.41] & $-$0.10 [$-$0.24, $+$0.06] \\
Flickr8k-Expert, all samples & 16,992 & $+$0.60 [$+$0.38, $+$0.81] & $+$0.47 [$+$0.26, $+$0.71] & $+$0.13 [$-$0.06, $+$0.30] \\
\addlinespace
\multicolumn{2}{l}{\emph{DENEB, Nebula training set}} & released $-$ dedup. & random $-$ dedup. & released $-$ random \\
Polaris test, all samples & 26,122 & $+$0.55 [$+$0.41, $+$0.70] & $+$0.36 [$+$0.25, $+$0.48] & $+$0.19 [$+$0.09, $+$0.28] \\
\quad duplicated & 13,111 & $+$0.92 [$+$0.65, $+$1.21] & $+$0.68 [$+$0.46, $+$0.92] & $+$0.24 [$+$0.07, $+$0.44] \\
\quad image only & 7,731 & $+$0.07 [$-$0.04, $+$0.18] & $+$0.03 [$-$0.06, $+$0.11] & $+$0.05 [$-$0.02, $+$0.11] \\
\quad unseen image & 5,280 & $+$0.59 [$+$0.27, $+$0.94] & $+$0.34 [$+$0.09, $+$0.59] & $+$0.25 [$+$0.02, $+$0.46] \\
\quad duplicated $-$ unseen & & $+$0.33 [$-$0.10, $+$0.76] & $+$0.34 [$+$0.02, $+$0.67] & $-$0.00 [$-$0.29, $+$0.28] \\
Spica test, all samples & 15,238 & $+$0.35 [$+$0.20, $+$0.53] & $+$0.21 [$+$0.09, $+$0.35] & $+$0.14 [$+$0.03, $+$0.25] \\
Nebula test, all samples & 3,298 & $+$0.58 [$+$0.35, $+$0.82] & $+$0.31 [$+$0.13, $+$0.48] & $+$0.28 [$+$0.12, $+$0.42] \\
Composite, all samples & 11,982 & $+$0.43 [$+$0.32, $+$0.55] & $+$0.18 [$+$0.08, $+$0.27] & $+$0.26 [$+$0.18, $+$0.34] \\
Flickr8k-Expert, all samples & 16,992 & $+$0.28 [$+$0.18, $+$0.39] & $+$0.26 [$+$0.14, $+$0.37] & $+$0.02 [$-$0.07, $+$0.10] \\
\bottomrule
\end{tabular}
}\end{table}

  \begin{table}[!htbp]\caption{Groups of Polaris test samples relative to the training set of each metric: size, share of tied sample pairs, $\tau$ of the deduplicated model, and differences between training sets, over all samples and above the lowest score.}\label{tab:groupsfull}\centering\apptablefont\makebox[\linewidth]{\begin{tabular}{lrrrrrr}
\toprule
Group of Polaris test samples & Samples & Images & Tied (\%) & $\tau$ dedup. & \multicolumn{2}{c}{Difference in $\tau$ [95\% CI]} \\
\midrule
\multicolumn{5}{l}{\emph{Pearl, Spica training set}} & released $-$ dedup. & matched $-$ dedup. \\
\quad duplicated & 17,856 & 2,371 & 29 & 28.2 & $+$6.45 [$+$5.84, $+$7.06] & $+$7.37 [$+$6.69, $+$8.05] \\
\quad duplicated, above lowest & 17,246 & 2,370 & 31 & 27.5 & $+$6.94 [$+$6.34, $+$7.59] & $+$7.92 [$+$7.23, $+$8.62] \\
\quad image only & 5,987 & 2,163 & 71 & 34.0 & $+$0.07 [$+$0.03, $+$0.13] & $+$0.08 [$+$0.02, $+$0.16] \\
\quad image only, above lowest & 958 & 136 & 29 & 29.1 & $+$2.64 [$+$1.05, $+$4.28] & $+$3.46 [$+$1.36, $+$5.93] \\
\quad unseen image & 2,279 & 243 & 23 & 59.4 & $+$1.20 [$+$0.19, $+$2.20] & $+$0.18 [$-$0.81, $+$1.08] \\
\quad unseen image, above lowest & 1,721 & 243 & 30 & 29.4 & $+$2.20 [$+$0.50, $+$3.93] & $+$0.20 [$-$1.51, $+$1.70] \\
\quad duplicated $-$ image only & & & & & $+$6.37 [$+$5.76, $+$6.99] & $+$7.28 [$+$6.59, $+$7.96] \\
\quad duplicated $-$ image only, above lowest & & & & & $+$4.30 [$+$2.51, $+$5.91] & $+$4.46 [$+$1.90, $+$6.62] \\
\quad duplicated $-$ unseen image & & & & & $+$5.24 [$+$4.06, $+$6.41] & $+$7.19 [$+$6.03, $+$8.37] \\
\quad duplicated $-$ unseen image, above lowest & & & & & $+$4.74 [$+$2.88, $+$6.58] & $+$7.72 [$+$6.08, $+$9.59] \\
\addlinespace
\multicolumn{5}{l}{\emph{DENEB, Nebula training set}} & released $-$ dedup. & random $-$ dedup. \\
\quad duplicated & 13,111 & 2,057 & 28 & 29.9 & $+$0.92 [$+$0.63, $+$1.20] & $+$0.68 [$+$0.47, $+$0.91] \\
\quad duplicated, above lowest & 12,659 & 2,055 & 30 & 30.3 & $+$0.91 [$+$0.62, $+$1.22] & $+$0.70 [$+$0.46, $+$0.94] \\
\quad image only & 7,731 & 1,948 & 40 & 63.6 & $+$0.07 [$-$0.04, $+$0.18] & $+$0.03 [$-$0.06, $+$0.11] \\
\quad image only, above lowest & 3,223 & 451 & 32 & 24.6 & $+$0.19 [$-$0.40, $+$0.78] & $+$0.07 [$-$0.32, $+$0.52] \\
\quad unseen image & 5,280 & 546 & 23 & 59.1 & $+$0.59 [$+$0.24, $+$0.90] & $+$0.34 [$+$0.10, $+$0.57] \\
\quad unseen image, above lowest & 4,043 & 545 & 30 & 31.1 & $+$0.84 [$+$0.29, $+$1.43] & $+$0.48 [$+$0.10, $+$0.91] \\
\quad duplicated $-$ image only & & & & & $+$0.85 [$+$0.55, $+$1.14] & $+$0.65 [$+$0.43, $+$0.89] \\
\quad duplicated $-$ image only, above lowest & & & & & $+$0.72 [$+$0.06, $+$1.35] & $+$0.63 [$+$0.10, $+$1.08] \\
\quad duplicated $-$ unseen image & & & & & $+$0.33 [$-$0.07, $+$0.78] & $+$0.34 [$+$0.04, $+$0.68] \\
\quad duplicated $-$ unseen image, above lowest & & & & & $+$0.07 [$-$0.57, $+$0.68] & $+$0.22 [$-$0.29, $+$0.66] \\
\bottomrule
\end{tabular}
}\end{table}

  \begin{table}[!htbp]\caption{Pearl after each training epoch: mean $100\times\tau$ over the trainings that reached the epoch (their number in parentheses).}\label{tab:epochs}\centering\apptablefont\makebox[\linewidth]{\begin{tabular}{lrrrrr}
\toprule
Training set & Epoch 1 & Epoch 2 & Epoch 3 & Epoch 4 & Epoch 5 \\
\midrule
\multicolumn{6}{l}{\emph{Polaris test, duplicated samples}} \\
\quad Released & 32.27\,(3) & 34.60\,(3) & 36.58\,(3) & -- & -- \\
\quad Random removal & 32.65\,(3) & 34.03\,(3) & 36.22\,(3) & 37.84\,(1) & -- \\
\quad Matched removal & 31.24\,(3) & 33.61\,(3) & 35.92\,(2) & 38.06\,(2) & 38.69\,(1) \\
\quad Deduplicated & 29.53\,(3) & 28.72\,(3) & 28.94\,(3) & 28.29\,(2) & 28.25\,(1) \\
\addlinespace
\multicolumn{6}{l}{\emph{Polaris test, all other samples}} \\
\quad Released & 57.11\,(3) & 57.21\,(3) & 57.23\,(3) & -- & -- \\
\quad Random removal & 57.13\,(3) & 57.01\,(3) & 57.20\,(3) & 57.18\,(1) & -- \\
\quad Matched removal & 56.94\,(3) & 57.05\,(3) & 57.09\,(2) & 57.13\,(2) & 57.14\,(1) \\
\quad Deduplicated & 56.92\,(3) & 56.90\,(3) & 57.00\,(3) & 56.88\,(2) & 57.00\,(1) \\
\addlinespace
\multicolumn{6}{l}{\emph{Composite, all samples}} \\
\quad Released & 62.80\,(3) & 62.29\,(3) & 61.65\,(3) & -- & -- \\
\quad Random removal & 62.32\,(3) & 62.18\,(3) & 62.03\,(3) & 61.40\,(1) & -- \\
\quad Matched removal & 62.13\,(3) & 62.31\,(3) & 61.89\,(2) & 61.75\,(2) & 61.77\,(1) \\
\quad Deduplicated & 61.88\,(3) & 61.20\,(3) & 61.28\,(3) & 61.03\,(2) & 60.86\,(1) \\
\addlinespace
\multicolumn{6}{l}{\emph{Flickr8k-Expert, all samples}} \\
\quad Released & 59.01\,(3) & 58.48\,(3) & 58.34\,(3) & -- & -- \\
\quad Random removal & 57.99\,(3) & 55.87\,(3) & 57.80\,(3) & 58.13\,(1) & -- \\
\quad Matched removal & 58.82\,(3) & 58.39\,(3) & 58.05\,(2) & 57.84\,(2) & 57.97\,(1) \\
\quad Deduplicated & 57.72\,(3) & 57.07\,(3) & 57.89\,(3) & 57.99\,(2) & 57.91\,(1) \\
\bottomrule
\end{tabular}
}\end{table}

  \begin{table}[!htbp]\caption{DENEB after each training epoch on the Polaris test set, as Table~\ref{tab:epochs}.}\label{tab:denebepochs}\centering\apptablefont\makebox[\linewidth]{\begin{tabular}{lrrrrrrrr}
\toprule
Training set & Epoch 1 & Epoch 2 & Epoch 3 & Epoch 4 & Epoch 5 & Epoch 6 & Epoch 7 & Epoch 8 \\
\midrule
\multicolumn{9}{l}{\emph{Duplicated samples}} \\
\quad With duplicates & 28.91\,(6) & 29.95\,(6) & 30.58\,(6) & 30.49\,(6) & 30.85\,(5) & 30.66\,(5) & 30.59\,(3) & 30.94\,(1) \\
\quad Deduplicated & 28.83\,(3) & 29.72\,(3) & 29.75\,(3) & 29.54\,(3) & 29.99\,(1) & -- & -- & -- \\
\addlinespace
\multicolumn{9}{l}{\emph{All other samples}} \\
\quad With duplicates & 67.64\,(6) & 68.09\,(6) & 68.29\,(6) & 68.30\,(6) & 68.42\,(5) & 68.41\,(5) & 68.41\,(3) & 68.62\,(1) \\
\quad Deduplicated & 67.65\,(3) & 68.08\,(3) & 68.20\,(3) & 68.17\,(3) & 68.27\,(1) & -- & -- & -- \\
\bottomrule
\end{tabular}
}\end{table}

\IfFileExists{figures/fig_epochs_deneb.pdf}{%
\begin{figure}[!htbp]
\centering
\includegraphics[width=\linewidth]{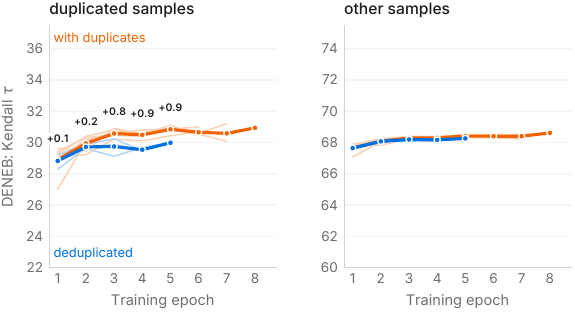}
\caption{DENEB on the Polaris test set after each training epoch, in the layout of Fig.~\ref{fig:epochs}: duplicated samples (left) and all other samples (right), relative to the Nebula training set.}
\label{fig:epochsdeneb}
\end{figure}}{}

Two patterns in Table~\ref{tab:epochs} go beyond the main text. On Composite all four training sets lose between half a point and one point from the first to the last epoch while the validation score does not fall, and the deduplicated set is the lowest at every epoch. On Flickr8k-Expert a single training can move by several points between consecutive epochs.

\ifdefined\DenebEpochGapLow
For DENEB we scored all \DenebEpochCheckpoints{} epoch checkpoints of the nine trainings (three seeds each on the released training set, after random removal and after deduplication; Table~\ref{tab:denebepochs}, Fig.~\ref{fig:epochsdeneb}). The deduplicated trainings stop after at most \DenebEpochCompared{} epochs, so the two groups can be compared up to that epoch. On duplicated samples the gap opens from \DenebEpochGapLow{} points after the first epoch to at most \DenebEpochGapHigh; on all other samples the two groups differ by at most \DenebEpochRestMax{} points. DENEB thus shows the same signature as Pearl at about a tenth of the size.
\fi

\section{Rigel by group of samples}
\label{app:rigel}

Table~\ref{tab:rigel} compares Rigel with the zero-shot InternVL judge, both with references, within each group of samples relative to the Spica training set. The last line applies the same comparison to Pearl trained on the released training set, for which the direct estimate is known (Section~\ref{sec:inflation}). Without references Rigel is \emph{below} the judge on the duplicated samples of the Polaris test set (\RigelFreePair).

  \begin{table}[!htbp]\caption{Lead of Rigel over a zero-shot judge, both with references, by group of samples (points of $\tau$, 95\% interval). Cells with fewer than 1{,}000 samples are omitted.}\label{tab:rigel}\centering\apptablefont\makebox[\linewidth]{\begin{tabular}{lrrrr}
\toprule
Rigel\,+\,ref $-$ judge\,+\,ref & All samples & Duplicated & Image only & Unseen image \\
\midrule
Polaris test & $+$1.5 [$+$1.1, $+$2.0] & $+$2.2 [$+$1.4, $+$2.9] & $+$0.1 [$-$0.0, $+$0.2] & $+$1.7 [$+$0.1, $+$3.2] \\
Spica test & $+$1.4 [$+$0.9, $+$1.9] & -- & $+$1.4 [$+$0.6, $+$2.3] & $+$1.4 [$+$0.7, $+$2.1] \\
Nebula test & $+$1.7 [$+$0.8, $+$2.4] & -- & $+$1.1 [$-$0.2, $+$2.5] & $+$2.1 [$+$1.1, $+$3.1] \\
\addlinespace
Pearl$^{\dagger}$, Polaris test & $+$0.8 [$+$0.3, $+$1.4] & $+$1.5 [$+$0.6, $+$2.5] & $-$0.1 [$-$0.2, $+$0.1] & $-$1.2 [$-$3.5, $+$0.9] \\
\bottomrule
\end{tabular}
}\end{table}

\section{Detection and ranking terms}
\label{app:components}

Table~\ref{tab:components} gives the weights of Eq.~\eqref{eq:components} for every evaluation set and the two abilities of the metrics. Flickr8k-CF is evaluated with $\tau_b$, for which the weights depend on the ties of the metric; the share of the detection term is computed there from the counts of concordant and discordant pairs directly. Fig.~\ref{fig:allabove} shows, for eleven metrics, how far $\tau$ falls when the lowest-score samples are set aside, next to the human reference.

  \begin{table}[!htbp]\caption{Detection and ranking by evaluation set: share of lowest-score samples, the weights of Eq.~\eqref{eq:components}, the best ranked metric with its $\tau$ over all samples, the share of that $\tau$ that comes from the detection term and its AUC, and the range of AUC and of $\tau$ above the lowest score over all ranked metrics other than text overlap.}\label{tab:components}\centering\apptablefont\makebox[\linewidth]{\begin{tabular}{lrrrlrrrrr}
\toprule
& Lowest & & & \multicolumn{4}{c}{Best ranked metric} & \multicolumn{2}{c}{Range, non-text metrics} \\
\cmidrule(lr){5-8} \cmidrule(l){9-10}
Evaluation set & (\%) & $w_{\text{det}}$ & $w_{\text{rank}}$ & name & $\tau$ & det. (\%) & AUC & AUC & $\tau(A)$ \\
\midrule
Polaris test & 24 & 0.45 & 0.55 & Judge InternVL-8B +\,ref & 60.5 & 70 & 0.967 & 0.96--0.97 & 20--33 \\
Spica test & 26 & 0.48 & 0.51 & Judge InternVL-8B +\,ref & 63.1 & 73 & 0.974 & 0.96--0.98 & 17--34 \\
Nebula test & 26 & 0.39 & 0.54 & Judge InternVL-8B +\,ref & 55.9 & 68 & 0.984 & 0.98--0.98 & 17--33 \\
Flickr8k-Expert & 58 & 0.65 & 0.16 & Rigel & 59.9 & 86 & 0.895 & 0.84--0.89 & 42--53 \\
Composite & 28 & 0.51 & 0.48 & Rigel +\,ref & 66.3 & 61 & 0.900 & 0.82--0.91 & 43--53 \\
Flickr8k-CF & 88 & -- & -- & Rigel +\,ref & 40.4 & 97 & 0.931 & 0.87--0.93 & 33--50 \\
\bottomrule
\end{tabular}
}\end{table}

\begin{figure}[!htbp]
\centering
\includegraphics[width=\linewidth]{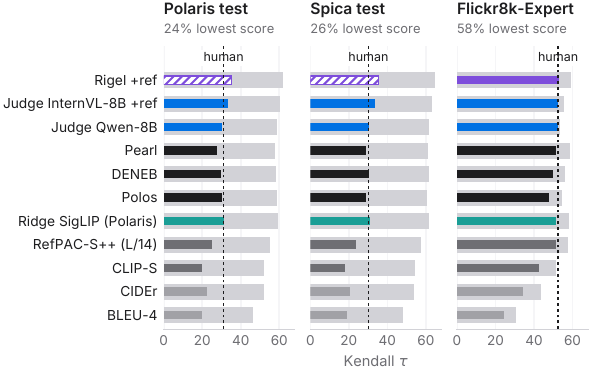}
\caption{Kendall's $\tau$ over all samples (wide grey bar) and above the lowest score (narrow coloured bar) for eleven metrics on the three sets with a human reference (dashed line, Eq.~\eqref{eq:human}). Hatched bars: the training set of the entry holds duplicated samples of that set, so it is not ranked there.}
\label{fig:allabove}
\end{figure}

\section{Full rankings}
\label{app:full}

Tables~\ref{tab:fullall}--\ref{tab:fullorder} give every metric on every set with its rank and the 95\% interval of the rank over bootstrap draws. A dagger marks a model whose training set holds duplicated samples of the set, a double dagger a model trained on the deduplicated training set on a set where the released one already qualifies; neither is ranked. Table~\ref{tab:unranked} compares the unranked entries with the best ranked metric, Table~\ref{tab:human} the best metrics with the human reference, Table~\ref{tab:learned} the three learned regression metrics with each other, and Table~\ref{tab:power} counts the separated pairs.

\begin{sidewaystable}[p]
\caption{$100\times$ Kendall's $\tau$ over all samples, with rank and 95\% rank interval.}
\label{tab:fullall}
\centering\smalltablefont\begin{tabular}{lrrrrrr}
\toprule
Metric & Polaris test & Spica test & Nebula test & Flickr8k-Expert & Composite & Flickr8k-CF \\
\midrule
Judge InternVL-8B +\,ref & 60.49\enspace1\,[1,\,2] & 63.11\enspace1\,[1,\,2] & 55.87\enspace1\,[1,\,2] & 55.43\enspace17\,[13,\,20] & 64.67\enspace4\,[3,\,5] & 40.14\enspace3\,[2,\,3] \\
Judge Qwen-8B +\,ref & 60.28\enspace2\,[1,\,3] & 62.79\enspace2\,[1,\,3] & 55.46\enspace2\,[1,\,3] & 53.76\enspace23\,[21,\,24] & 65.62\enspace2\,[2,\,2] & 39.90\enspace4\,[4,\,6] \\
Judge Qwen-4B +\,ref & 60.08\enspace3\,[2,\,3] & 62.46\enspace3\,[2,\,4] & 54.85\enspace3\,[2,\,7] & 55.39\enspace18\,[14,\,20] & 64.54\enspace5\,[3,\,6] & 39.88\enspace5\,[4,\,6] \\
Ridge SigLIP (Polaris) & 59.45\enspace4\,[4,\,6] & 61.82\enspace4\,[3,\,7] & 54.69\enspace4\,[2,\,7] & 58.08\enspace5\,[4,\,6] & 59.39\enspace14\,[13,\,16] & 39.01\enspace10\,[10,\,11] \\
Judge Qwen-4B & 58.97\enspace5\,[4,\,8] & 61.64\enspace6\,[4,\,8] & 53.78\enspace9\,[5,\,11] & 55.31\enspace19\,[15,\,20] & 61.99\enspace10\,[8,\,12] & 39.74\enspace6\,[5,\,8] \\
Judge Qwen-8B & 58.91\enspace6\,[4,\,8] & 61.65\enspace5\,[4,\,8] & 54.49\enspace5\,[3,\,8] & 53.44\enspace24\,[23,\,25] & 64.76\enspace3\,[3,\,5] & 39.73\enspace7\,[5,\,8] \\
Polos & 58.89\enspace7\,[5,\,8] & 60.57\enspace10\,[8,\,12] & 52.61\enspace12\,[9,\,13] & 54.67\enspace20\,[18,\,22] & 58.81\enspace17\,[16,\,19] & 37.00\enspace22\,[21,\,23] \\
Judge InternVL-8B & 58.57\enspace8\,[6,\,11] & 61.11\enspace8\,[6,\,10] & 53.39\enspace10\,[7,\,12] & 55.88\enspace13\,[11,\,17] & 62.48\enspace7\,[7,\,10] & 39.53\enspace9\,[7,\,9] \\
Ridge CLIP (Polaris) & 58.29\enspace9\,[7,\,11] & 60.12\enspace12\,[10,\,12] & 52.83\enspace11\,[9,\,12] & 56.97\enspace9\,[8,\,10] & 57.20\enspace23\,[21,\,23] & 37.72\enspace19\,[17,\,20] \\
DENEB & 58.26\enspace10\,[7,\,11] & 61.51\enspace7\,[4,\,9] & 54.02\enspace7\,[4,\,10] & 56.34\enspace12\,[11,\,14] & 59.58\enspace13\,[13,\,15] & 38.07\enspace15\,[15,\,17] \\
Polos (official) & 57.87\enspace11\,[10,\,12] & -- & -- & 56.45\enspace11\,[10,\,14] & 58.13\enspace20\,[18,\,21] & 37.80\enspace18\,[16,\,20] \\
Pearl & 57.61\enspace12\,[11,\,13] & 61.02\enspace9\,[6,\,10] & 54.25\enspace6\,[3,\,10] & 58.48\enspace3\,[2,\,5] & 62.29\enspace8\,[7,\,10] & 38.81\enspace12\,[11,\,12] \\
Ridge SigLIP (Spica) & 57.11\enspace13\,[12,\,13] & 60.21\enspace11\,[10,\,12] & 53.84\enspace8\,[5,\,10] & 58.45\enspace4\,[3,\,4] & 62.10\enspace9\,[7,\,11] & 39.70\enspace8\,[4,\,9] \\
Ridge CLIP (Spica) & 56.23\enspace14\,[14,\,14] & 58.48\enspace13\,[13,\,14] & 52.04\enspace13\,[12,\,13] & 57.39\enspace8\,[6,\,9] & 59.34\enspace15\,[13,\,16] & 38.35\enspace14\,[13,\,15] \\
RefPAC-S++ (L/14) & 55.36\enspace15\,[15,\,18] & 57.61\enspace15\,[14,\,16] & 50.81\enspace14\,[14,\,16] & 57.91\enspace6\,[4,\,7] & 61.59\enspace12\,[11,\,12] & 38.83\enspace11\,[10,\,12] \\
RefPAC-S (L/14) & 55.33\enspace16\,[15,\,18] & 57.71\enspace14\,[13,\,16] & 50.43\enspace16\,[14,\,18] & 56.88\enspace10\,[8,\,11] & 58.61\enspace18\,[16,\,19] & 37.99\enspace16\,[15,\,19] \\
RefPAC-S & 55.20\enspace17\,[15,\,18] & 57.49\enspace16\,[14,\,16] & 50.65\enspace15\,[14,\,17] & 55.86\enspace14\,[12,\,18] & 57.32\enspace22\,[21,\,23] & 37.62\enspace20\,[18,\,20] \\
RefPAC-S++ & 54.83\enspace18\,[17,\,19] & 56.37\enspace18\,[17,\,19] & 49.84\enspace17\,[16,\,19] & 55.76\enspace15\,[13,\,19] & 59.03\enspace16\,[14,\,17] & 37.98\enspace17\,[15,\,18] \\
RefCLIP-S & 54.76\enspace19\,[16,\,19] & 56.76\enspace17\,[16,\,18] & 49.78\enspace18\,[15,\,19] & 53.01\enspace25\,[23,\,25] & 55.41\enspace25\,[24,\,26] & 36.40\enspace24\,[24,\,24] \\
Text-only SimCSE & 53.79\enspace20\,[20,\,22] & 55.22\enspace21\,[19,\,23] & 49.10\enspace19\,[17,\,21] & 49.86\enspace27\,[27,\,27] & 55.11\enspace26\,[24,\,26] & 34.68\enspace26\,[26,\,27] \\
PAC-S++ (L/14) & 53.57\enspace21\,[20,\,22] & 55.81\enspace19\,[18,\,21] & 48.86\enspace20\,[18,\,21] & 57.40\enspace7\,[7,\,9] & 61.98\enspace11\,[8,\,11] & 38.49\enspace13\,[13,\,14] \\
PAC-S (L/14) & 53.12\enspace22\,[21,\,23] & 55.40\enspace20\,[19,\,21] & 47.92\enspace22\,[20,\,24] & 55.71\enspace16\,[13,\,19] & 57.75\enspace21\,[20,\,22] & 37.02\enspace21\,[21,\,23] \\
PAC-S++ & 52.36\enspace23\,[23,\,26] & 53.73\enspace24\,[23,\,27] & 46.91\enspace25\,[22,\,27] & 54.49\enspace21\,[19,\,22] & 58.29\enspace19\,[17,\,20] & 36.97\enspace23\,[21,\,23] \\
PAC-S & 52.32\enspace24\,[23,\,26] & 54.41\enspace22\,[21,\,25] & 47.18\enspace23\,[21,\,27] & 54.29\enspace22\,[20,\,24] & 55.72\enspace24\,[24,\,26] & 36.04\enspace25\,[25,\,25] \\
CLIP-S & 52.30\enspace25\,[22,\,26] & 54.15\enspace23\,[21,\,26] & 46.88\enspace26\,[22,\,27] & 51.44\enspace26\,[26,\,26] & 53.82\enspace27\,[27,\,27] & 34.34\enspace27\,[26,\,27] \\
CIDEr & 52.06\enspace26\,[23,\,26] & 53.60\enspace25\,[22,\,26] & 48.07\enspace21\,[20,\,25] & 43.89\enspace29\,[28,\,29] & 37.73\enspace30\,[30,\,30] & 24.62\enspace28\,[28,\,29] \\
METEOR & 51.22\enspace27\,[26,\,28] & 52.85\enspace27\,[25,\,27] & 46.79\enspace27\,[22,\,27] & 41.82\enspace30\,[30,\,30] & 38.98\enspace29\,[29,\,29] & 22.27\enspace30\,[30,\,30] \\
SPICE & 50.94\enspace28\,[27,\,28] & 53.43\enspace26\,[22,\,27] & 46.93\enspace24\,[22,\,27] & 44.89\enspace28\,[28,\,29] & 40.40\enspace28\,[28,\,28] & 24.48\enspace29\,[28,\,29] \\
ROUGE-L & 46.24\enspace29\,[29,\,30] & 47.28\enspace29\,[28,\,29] & 42.58\enspace29\,[28,\,29] & 32.31\enspace32\,[31,\,32] & 32.52\enspace31\,[31,\,31] & 19.90\enspace31\,[31,\,31] \\
BLEU-4 & 46.23\enspace30\,[29,\,30] & 48.05\enspace28\,[28,\,29] & 43.46\enspace28\,[28,\,29] & 30.76\enspace33\,[33,\,33] & 30.64\enspace33\,[32,\,33] & 16.88\enspace33\,[33,\,33] \\
BLEU-1 & 45.26\enspace31\,[31,\,31] & 46.05\enspace30\,[30,\,30] & 40.20\enspace30\,[30,\,30] & 32.35\enspace31\,[31,\,32] & 31.31\enspace32\,[32,\,33] & 18.11\enspace32\,[32,\,32] \\
Rigel & -- & -- & -- & 59.88\enspace1\,[1,\,1] & 64.08\enspace6\,[4,\,6] & 40.36\enspace2\,[1,\,3] \\
Rigel +\,ref & -- & -- & -- & 59.01\enspace2\,[2,\,3] & 66.28\enspace1\,[1,\,1] & 40.42\enspace1\,[1,\,2] \\
\addlinespace
Rigel$^{\dagger}$ & 57.85 & 61.14 & 55.02 & -- & -- & -- \\
Rigel +\,ref$^{\dagger}$ & 62.01 & 64.52 & 57.52 & -- & -- & -- \\
DENEB$^{\dagger}$ & 58.81 & -- & -- & -- & -- & -- \\
Pearl$^{\dagger}$ & 61.32 & 62.09 & 54.91 & -- & -- & -- \\
Ridge SigLIP (Spica)$^{\dagger}$ & 58.90 & 61.09 & 54.46 & -- & -- & -- \\
Ridge CLIP (Spica)$^{\dagger}$ & 58.02 & 59.27 & 52.69 & -- & -- & -- \\
Polos (official)$^{\dagger}$ & -- & 60.49 & 53.67 & -- & -- & -- \\
Polos$^{\dagger}$ & -- & 63.06 & 54.31 & -- & -- & -- \\
Ridge SigLIP (Polaris)$^{\dagger}$ & -- & 63.02 & 55.55 & -- & -- & -- \\
Ridge CLIP (Polaris)$^{\dagger}$ & -- & 61.41 & 53.63 & -- & -- & -- \\
\addlinespace
Polos$^{\ddagger}$ & 58.66 & -- & -- & 54.89 & 58.04 & 37.25 \\
Ridge SigLIP (Polaris)$^{\ddagger}$ & 59.32 & -- & -- & 58.32 & 59.26 & 39.03 \\
Ridge CLIP (Polaris)$^{\ddagger}$ & 58.19 & -- & -- & 57.15 & 56.68 & 37.70 \\
DENEB$^{\ddagger}$ & -- & 61.16 & 53.44 & 56.07 & 59.15 & 37.80 \\
Pearl$^{\ddagger}$ & -- & -- & -- & 57.88 & 61.18 & 38.68 \\
Ridge SigLIP (Spica)$^{\ddagger}$ & -- & -- & -- & 58.90 & 62.16 & 39.69 \\
Ridge CLIP (Spica)$^{\ddagger}$ & -- & -- & -- & 57.68 & 59.23 & 38.36 \\
\bottomrule
\end{tabular}

\end{sidewaystable}

\begin{sidewaystable}[p]
\caption{$100\times$ Kendall's $\tau$ above the lowest score, with rank and 95\% rank interval.}
\label{tab:fullabove}
\centering\smalltablefont\begin{tabular}{lrrrrrr}
\toprule
Metric & Polaris test & Spica test & Nebula test & Flickr8k-Expert & Composite & Flickr8k-CF \\
\midrule
Judge InternVL-8B +\,ref & 33.43\enspace1\,[1,\,2] & 33.75\enspace1\,[1,\,2] & 33.02\enspace1\,[1,\,2] & 52.71\enspace4\,[1,\,6] & 52.26\enspace3\,[3,\,4] & 48.63\enspace2\,[2,\,3] \\
Judge Qwen-8B +\,ref & 32.82\enspace2\,[2,\,3] & 33.03\enspace2\,[2,\,3] & 32.40\enspace2\,[1,\,3] & 52.26\enspace5\,[3,\,9] & 52.85\enspace2\,[1,\,2] & 47.66\enspace4\,[3,\,5] \\
Judge Qwen-4B +\,ref & 32.78\enspace3\,[2,\,3] & 32.86\enspace3\,[2,\,3] & 31.25\enspace3\,[3,\,7] & 51.67\enspace7\,[6,\,13] & 51.81\enspace4\,[3,\,5] & 45.18\enspace7\,[6,\,9] \\
Ridge SigLIP (Polaris) & 31.42\enspace4\,[4,\,5] & 31.05\enspace5\,[4,\,7] & 30.86\enspace4\,[2,\,7] & 51.52\enspace10\,[5,\,13] & 49.17\enspace12\,[10,\,12] & 45.05\enspace8\,[6,\,10] \\
Judge Qwen-4B & 30.64\enspace5\,[4,\,9] & 31.32\enspace4\,[4,\,7] & 29.51\enspace8\,[5,\,11] & 51.36\enspace11\,[7,\,14] & 49.26\enspace11\,[9,\,12] & 43.69\enspace10\,[9,\,14] \\
Polos & 30.47\enspace6\,[4,\,9] & 28.97\enspace10\,[8,\,11] & 27.51\enspace12\,[9,\,13] & 48.02\enspace21\,[21,\,24] & 47.39\enspace19\,[13,\,20] & 38.01\enspace25\,[22,\,25] \\
Judge Qwen-8B & 30.13\enspace7\,[5,\,10] & 30.60\enspace6\,[4,\,8] & 30.60\enspace5\,[3,\,8] & 52.21\enspace6\,[3,\,10] & 51.63\enspace5\,[3,\,6] & 45.72\enspace6\,[6,\,8] \\
Judge InternVL-8B & 29.98\enspace8\,[5,\,11] & 29.86\enspace8\,[6,\,10] & 28.72\enspace10\,[7,\,12] & 51.63\enspace8\,[6,\,13] & 49.65\enspace10\,[7,\,11] & 44.92\enspace9\,[6,\,10] \\
DENEB & 29.75\enspace9\,[6,\,11] & 30.47\enspace7\,[4,\,8] & 29.73\enspace7\,[4,\,10] & 49.89\enspace17\,[14,\,20] & 47.85\enspace13\,[13,\,18] & 42.86\enspace12\,[10,\,15] \\
Ridge CLIP (Polaris) & 29.64\enspace10\,[6,\,10] & 28.28\enspace11\,[9,\,12] & 27.69\enspace11\,[9,\,12] & 49.62\enspace18\,[14,\,20] & 46.71\enspace21\,[19,\,21] & 41.01\enspace19\,[16,\,20] \\
Polos (official) & 28.93\enspace11\,[9,\,11] & -- & -- & 49.44\enspace19\,[15,\,20] & 46.10\enspace22\,[22,\,22] & 41.43\enspace18\,[15,\,19] \\
Pearl & 27.78\enspace12\,[12,\,13] & 29.05\enspace9\,[7,\,11] & 30.17\enspace6\,[3,\,10] & 51.35\enspace12\,[6,\,13] & 50.18\enspace7\,[6,\,10] & 42.06\enspace16\,[12,\,18] \\
Ridge SigLIP (Spica) & 26.92\enspace13\,[12,\,13] & 27.66\enspace12\,[10,\,12] & 29.32\enspace9\,[5,\,11] & 52.89\enspace3\,[1,\,6] & 50.00\enspace8\,[6,\,11] & 47.21\enspace5\,[3,\,5] \\
Ridge CLIP (Spica) & 25.59\enspace14\,[14,\,17] & 24.83\enspace13\,[13,\,15] & 26.21\enspace13\,[12,\,13] & 51.05\enspace13\,[7,\,14] & 46.92\enspace20\,[17,\,21] & 42.37\enspace14\,[11,\,17] \\
RefPAC-S++ (L/14) & 24.84\enspace15\,[14,\,19] & 23.55\enspace15\,[13,\,19] & 23.57\enspace15\,[14,\,19] & 51.54\enspace9\,[5,\,13] & 49.97\enspace9\,[7,\,11] & 43.56\enspace11\,[9,\,12] \\
RefPAC-S (L/14) & 24.82\enspace16\,[14,\,19] & 23.62\enspace14\,[13,\,18] & 23.11\enspace16\,[14,\,21] & 49.94\enspace16\,[14,\,19] & 47.83\enspace14\,[13,\,18] & 42.42\enspace13\,[12,\,17] \\
BLEU-1 & 24.80\enspace17\,[14,\,20] & 22.81\enspace17\,[13,\,21] & 20.74\enspace23\,[17,\,26] & 27.87\enspace31\,[30,\,32] & 28.96\enspace32\,[32,\,32] & 23.06\enspace31\,[31,\,31] \\
RefPAC-S & 24.48\enspace18\,[15,\,20] & 23.34\enspace16\,[14,\,19] & 23.58\enspace14\,[14,\,19] & 49.16\enspace20\,[17,\,20] & 45.25\enspace23\,[23,\,24] & 40.77\enspace20\,[17,\,21] \\
METEOR & 23.89\enspace19\,[16,\,22] & 22.40\enspace19\,[14,\,21] & 21.86\enspace20\,[15,\,23] & 29.47\enspace30\,[29,\,31] & 34.78\enspace28\,[28,\,28] & 25.10\enspace29\,[29,\,30] \\
RefPAC-S++ & 23.82\enspace20\,[18,\,22] & 21.55\enspace21\,[18,\,24] & 21.95\enspace19\,[16,\,24] & 50.03\enspace15\,[14,\,19] & 47.49\enspace18\,[14,\,20] & 42.08\enspace15\,[12,\,18] \\
RefCLIP-S & 23.73\enspace21\,[17,\,22] & 22.45\enspace18\,[15,\,21] & 22.41\enspace17\,[15,\,24] & 47.09\enspace24\,[22,\,24] & 43.60\enspace25\,[25,\,26] & 39.37\enspace21\,[21,\,23] \\
Text-only SimCSE & 23.24\enspace22\,[19,\,23] & 20.54\enspace22\,[20,\,27] & 21.61\enspace22\,[16,\,25] & 45.62\enspace25\,[24,\,26] & 41.73\enspace27\,[27,\,27] & 39.30\enspace22\,[20,\,24] \\
CIDEr & 22.34\enspace23\,[21,\,26] & 20.35\enspace23\,[21,\,27] & 22.06\enspace18\,[14,\,23] & 34.17\enspace28\,[28,\,28] & 30.81\enspace30\,[30,\,31] & 28.05\enspace28\,[28,\,28] \\
PAC-S++ (L/14) & 21.78\enspace24\,[22,\,27] & 20.26\enspace24\,[20,\,27] & 20.00\enspace25\,[19,\,27] & 50.40\enspace14\,[11,\,18] & 50.73\enspace6\,[5,\,8] & 41.44\enspace17\,[14,\,20] \\
SPICE & 21.59\enspace25\,[23,\,28] & 21.80\enspace20\,[15,\,23] & 21.71\enspace21\,[14,\,25] & 30.22\enspace29\,[29,\,31] & 32.22\enspace29\,[29,\,29] & 19.87\enspace33\,[32,\,33] \\
ROUGE-L & 21.48\enspace26\,[23,\,29] & 19.93\enspace25\,[21,\,28] & 20.50\enspace24\,[20,\,27] & 26.70\enspace32\,[31,\,32] & 30.36\enspace31\,[30,\,31] & 25.07\enspace30\,[29,\,30] \\
PAC-S (L/14) & 21.03\enspace27\,[24,\,29] & 19.44\enspace26\,[22,\,28] & 18.61\enspace27\,[24,\,29] & 47.56\enspace22\,[21,\,24] & 47.53\enspace16\,[13,\,21] & 38.63\enspace23\,[21,\,25] \\
BLEU-4 & 19.94\enspace28\,[26,\,31] & 19.19\enspace27\,[23,\,30] & 19.53\enspace26\,[23,\,29] & 24.42\enspace33\,[33,\,33] & 28.05\enspace33\,[33,\,33] & 20.94\enspace32\,[32,\,33] \\
CLIP-S & 19.82\enspace29\,[27,\,31] & 17.98\enspace28\,[25,\,30] & 17.51\enspace28\,[26,\,30] & 42.44\enspace27\,[27,\,27] & 43.35\enspace26\,[25,\,26] & 33.10\enspace27\,[27,\,27] \\
PAC-S & 19.69\enspace30\,[27,\,31] & 17.95\enspace29\,[26,\,30] & 17.40\enspace29\,[26,\,30] & 45.49\enspace26\,[25,\,26] & 45.01\enspace24\,[23,\,24] & 35.05\enspace26\,[26,\,26] \\
PAC-S++ & 19.62\enspace31\,[27,\,31] & 16.86\enspace30\,[28,\,30] & 16.66\enspace30\,[28,\,30] & 47.47\enspace23\,[21,\,24] & 47.52\enspace17\,[13,\,21] & 38.18\enspace24\,[22,\,25] \\
Rigel & -- & -- & -- & 53.32\enspace1\,[1,\,3] & 47.60\enspace15\,[13,\,21] & 48.05\enspace3\,[2,\,5] \\
Rigel +\,ref & -- & -- & -- & 53.01\enspace2\,[1,\,5] & 53.39\enspace1\,[1,\,2] & 49.92\enspace1\,[1,\,1] \\
\addlinespace
Rigel$^{\dagger}$ & 27.71 & 29.16 & 31.47 & -- & -- & -- \\
Rigel +\,ref$^{\dagger}$ & 35.19 & 35.55 & 35.73 & -- & -- & -- \\
DENEB$^{\dagger}$ & 30.54 & -- & -- & -- & -- & -- \\
Pearl$^{\dagger}$ & 34.12 & 31.10 & 31.18 & -- & -- & -- \\
Ridge SigLIP (Spica)$^{\dagger}$ & 29.97 & 29.35 & 30.45 & -- & -- & -- \\
Ridge CLIP (Spica)$^{\dagger}$ & 28.64 & 26.35 & 27.38 & -- & -- & -- \\
Polos (official)$^{\dagger}$ & -- & 28.85 & 29.29 & -- & -- & -- \\
Polos$^{\dagger}$ & -- & 33.28 & 30.54 & -- & -- & -- \\
Ridge SigLIP (Polaris)$^{\dagger}$ & -- & 33.17 & 32.45 & -- & -- & -- \\
Ridge CLIP (Polaris)$^{\dagger}$ & -- & 30.51 & 29.15 & -- & -- & -- \\
\addlinespace
Polos$^{\ddagger}$ & 30.22 & -- & -- & 48.18 & 46.74 & 38.81 \\
Ridge SigLIP (Polaris)$^{\ddagger}$ & 31.20 & -- & -- & 51.71 & 48.89 & 45.14 \\
Ridge CLIP (Polaris)$^{\ddagger}$ & 29.49 & -- & -- & 49.70 & 46.21 & 41.02 \\
DENEB$^{\ddagger}$ & -- & 29.96 & 28.70 & 49.09 & 47.82 & 41.77 \\
Pearl$^{\ddagger}$ & -- & -- & -- & 51.45 & 48.40 & 42.18 \\
Ridge SigLIP (Spica)$^{\ddagger}$ & -- & -- & -- & 53.17 & 49.63 & 46.87 \\
Ridge CLIP (Spica)$^{\ddagger}$ & -- & -- & -- & 51.32 & 46.45 & 42.06 \\
\bottomrule
\end{tabular}

\end{sidewaystable}

\begin{sidewaystable}[p]
\caption{Order accuracy (\%) for two candidates of one image (Eq.~\eqref{eq:order}), with rank and 95\% rank interval; the human reference is the other half of the ratings.}
\label{tab:fullorder}
\centering\smalltablefont\begin{tabular}{lrrr}
\toprule
Metric & Polaris test & Spica test & Flickr8k-Expert \\
\midrule
Judge InternVL-8B +\,ref & 72.83\enspace1\,[1,\,6] & 73.14\enspace1\,[1,\,5] & 84.28\enspace13\,[7,\,18] \\
Judge Qwen-8B +\,ref & 72.37\enspace2\,[1,\,7] & 72.63\enspace2\,[1,\,6] & 83.22\enspace20\,[17,\,23] \\
Judge Qwen-4B +\,ref & 72.06\enspace3\,[1,\,7] & 72.07\enspace3\,[1,\,7] & 84.38\enspace12\,[7,\,17] \\
Polos & 72.04\enspace4\,[1,\,8] & 69.98\enspace7\,[2,\,11] & 83.05\enspace21\,[18,\,24] \\
Ridge SigLIP (Polaris) & 70.79\enspace5\,[2,\,11] & 69.66\enspace9\,[3,\,11] & 85.44\enspace4\,[4,\,7] \\
Ridge CLIP (Polaris) & 70.62\enspace6\,[2,\,12] & 67.16\enspace11\,[8,\,19] & 84.17\enspace15\,[9,\,18] \\
Judge Qwen-8B & 70.58\enspace7\,[3,\,13] & 70.78\enspace5\,[1,\,10] & 82.72\enspace24\,[20,\,24] \\
DENEB & 70.52\enspace8\,[3,\,12] & 69.79\enspace8\,[2,\,11] & 84.48\enspace10\,[7,\,15] \\
Pearl & 69.48\enspace9\,[5,\,15] & 68.07\enspace10\,[6,\,15] & 85.32\enspace5\,[4,\,7] \\
Judge InternVL-8B & 69.35\enspace10\,[5,\,17] & 70.88\enspace4\,[2,\,11] & 84.38\enspace11\,[7,\,17] \\
Judge Qwen-4B & 69.35\enspace11\,[5,\,18] & 70.54\enspace6\,[2,\,11] & 83.75\enspace18\,[13,\,21] \\
Polos (official) & 69.33\enspace12\,[5,\,16] & -- & 83.92\enspace16\,[12,\,19] \\
Ridge SigLIP (Spica) & 68.11\enspace13\,[9,\,20] & 65.63\enspace13\,[10,\,26] & 85.98\enspace3\,[2,\,3] \\
RefPAC-S (L/14) & 67.85\enspace14\,[9,\,20] & 64.89\enspace15\,[10,\,27] & 84.71\enspace9\,[6,\,13] \\
Ridge CLIP (Spica) & 67.68\enspace15\,[10,\,21] & 64.25\enspace19\,[12,\,29] & 84.75\enspace8\,[5,\,13] \\
PAC-S (L/14) & 67.58\enspace16\,[9,\,22] & 62.83\enspace25\,[14,\,30] & 84.27\enspace14\,[9,\,18] \\
CIDEr & 66.61\enspace17\,[10,\,26] & 63.75\enspace22\,[12,\,30] & 74.58\enspace29\,[28,\,30] \\
RefPAC-S & 66.50\enspace18\,[13,\,25] & 65.53\enspace14\,[10,\,25] & 83.87\enspace17\,[12,\,19] \\
METEOR & 66.14\enspace19\,[12,\,27] & 64.68\enspace17\,[10,\,28] & 74.10\enspace30\,[29,\,30] \\
PAC-S++ (L/14) & 65.72\enspace20\,[15,\,28] & 63.78\enspace21\,[12,\,30] & 84.81\enspace7\,[6,\,13] \\
RefPAC-S++ (L/14) & 65.37\enspace21\,[16,\,29] & 65.70\enspace12\,[10,\,24] & 85.08\enspace6\,[4,\,9] \\
RefCLIP-S & 64.95\enspace22\,[17,\,30] & 63.71\enspace23\,[14,\,28] & 81.62\enspace25\,[25,\,26] \\
BLEU-1 & 64.84\enspace23\,[16,\,30] & 62.74\enspace27\,[14,\,30] & 69.32\enspace31\,[31,\,32] \\
Text-only SimCSE & 64.75\enspace24\,[16,\,31] & 64.05\enspace20\,[11,\,30] & 79.99\enspace27\,[27,\,27] \\
RefPAC-S++ & 64.74\enspace25\,[18,\,30] & 62.35\enspace29\,[18,\,30] & 83.45\enspace19\,[16,\,22] \\
ROUGE-L & 64.13\enspace26\,[19,\,31] & 61.81\enspace30\,[17,\,30] & 68.86\enspace32\,[31,\,33] \\
CLIP-S & 64.04\enspace27\,[18,\,31] & 62.75\enspace26\,[13,\,30] & 81.14\enspace26\,[25,\,26] \\
PAC-S & 63.89\enspace28\,[19,\,31] & 64.50\enspace18\,[10,\,29] & 82.91\enspace23\,[19,\,24] \\
BLEU-4 & 63.63\enspace29\,[19,\,31] & 63.20\enspace24\,[12,\,30] & 67.78\enspace33\,[32,\,33] \\
PAC-S++ & 63.57\enspace30\,[21,\,31] & 62.66\enspace28\,[14,\,30] & 82.98\enspace22\,[19,\,24] \\
SPICE & 62.46\enspace31\,[24,\,31] & 64.76\enspace16\,[10,\,28] & 75.37\enspace28\,[28,\,29] \\
Rigel & -- & -- & 86.63\enspace1\,[1,\,2] \\
Rigel +\,ref & -- & -- & 86.52\enspace2\,[1,\,3] \\
\addlinespace
Rigel$^{\dagger}$ & 67.97 & 68.55 & -- \\
Rigel +\,ref$^{\dagger}$ & 72.59 & 73.58 & -- \\
DENEB$^{\dagger}$ & 70.44 & -- & -- \\
Pearl$^{\dagger}$ & 72.25 & 71.08 & -- \\
Ridge SigLIP (Spica)$^{\dagger}$ & 69.21 & 67.51 & -- \\
Ridge CLIP (Spica)$^{\dagger}$ & 68.61 & 66.02 & -- \\
Polos (official)$^{\dagger}$ & -- & 66.93 & -- \\
Polos$^{\dagger}$ & -- & 72.68 & -- \\
Ridge SigLIP (Polaris)$^{\dagger}$ & -- & 70.65 & -- \\
Ridge CLIP (Polaris)$^{\dagger}$ & -- & 68.84 & -- \\
\addlinespace
Polos$^{\ddagger}$ & 71.67 & -- & 83.40 \\
Ridge SigLIP (Polaris)$^{\ddagger}$ & 70.58 & -- & 85.72 \\
Ridge CLIP (Polaris)$^{\ddagger}$ & 70.48 & -- & 84.21 \\
DENEB$^{\ddagger}$ & -- & 69.88 & 84.15 \\
Pearl$^{\ddagger}$ & -- & -- & 84.55 \\
Ridge SigLIP (Spica)$^{\ddagger}$ & -- & -- & 86.05 \\
Ridge CLIP (Spica)$^{\ddagger}$ & -- & -- & 84.73 \\
\midrule
Human reference & 66.38 & 68.52 & 77.67 \\
\bottomrule
\end{tabular}

\end{sidewaystable}

  \begin{table}[!htbp]\caption{Entries that are not ranked ($\tau$ over all samples): difference to the best ranked metric, share of bootstrap draws in which the entry is above every ranked metric, and difference to the ranked model of the same metric.}\label{tab:unranked}\centering\apptablefont\makebox[\linewidth]{\begin{tabular}{llrrrr}
\toprule
Set & Entry & $\tau$ & $-$ best ranked & first (\%) & $-$ own ranked version \\
\midrule
Polaris test & Polos$^{\ddagger}$ & 58.66 & $-$1.84 [$-$2.57, $-$1.12] & 0.0 & $-$0.23 [$-$0.52, $+$0.06] \\
 & Rigel$^{\dagger}$ & 57.85 & $-$2.64 [$-$3.33, $-$1.95] & 0.0 & -- \\
 & Rigel +\,ref$^{\dagger}$ & 62.01 & $+$1.52 [$+$1.07, $+$1.97] & 100.0 & -- \\
 & DENEB$^{\dagger}$ & 58.81 & $-$1.68 [$-$2.28, $-$1.10] & 0.0 & $+$0.55 [$+$0.40, $+$0.69] \\
 & Pearl$^{\dagger}$ & 61.32 & $+$0.83 [$+$0.24, $+$1.40] & 99.8 & $+$3.72 [$+$3.39, $+$4.05] \\
 & Ridge SigLIP (Polaris)$^{\ddagger}$ & 59.32 & $-$1.17 [$-$1.77, $-$0.58] & 0.0 & $-$0.13 [$-$0.22, $-$0.04] \\
 & Ridge SigLIP (Spica)$^{\dagger}$ & 58.90 & $-$1.59 [$-$2.22, $-$0.96] & 0.0 & $+$1.80 [$+$1.63, $+$1.97] \\
 & Ridge CLIP (Polaris)$^{\ddagger}$ & 58.19 & $-$2.30 [$-$2.99, $-$1.64] & 0.0 & $-$0.10 [$-$0.22, $+$0.02] \\
 & Ridge CLIP (Spica)$^{\dagger}$ & 58.02 & $-$2.47 [$-$3.18, $-$1.78] & 0.0 & $+$1.80 [$+$1.61, $+$1.99] \\
\addlinespace
Spica test & Polos (official)$^{\dagger}$ & 60.49 & $-$2.62 [$-$3.46, $-$1.83] & 0.0 & -- \\
 & Polos$^{\dagger}$ & 63.06 & $-$0.05 [$-$0.79, $+$0.66] & 43.9 & $+$2.48 [$+$2.07, $+$2.90] \\
 & Rigel$^{\dagger}$ & 61.14 & $-$1.97 [$-$2.78, $-$1.15] & 0.0 & -- \\
 & Rigel +\,ref$^{\dagger}$ & 64.52 & $+$1.41 [$+$0.87, $+$1.94] & 100.0 & -- \\
 & DENEB$^{\ddagger}$ & 61.16 & $-$1.95 [$-$2.73, $-$1.23] & 0.0 & $-$0.35 [$-$0.52, $-$0.18] \\
 & Pearl$^{\dagger}$ & 62.09 & $-$1.02 [$-$1.74, $-$0.32] & 0.2 & $+$1.08 [$+$0.77, $+$1.39] \\
 & Ridge SigLIP (Polaris)$^{\dagger}$ & 63.02 & $-$0.09 [$-$0.80, $+$0.57] & 39.6 & $+$1.20 [$+$1.05, $+$1.36] \\
 & Ridge SigLIP (Spica)$^{\dagger}$ & 61.09 & $-$2.02 [$-$2.82, $-$1.26] & 0.0 & $+$0.88 [$+$0.73, $+$1.04] \\
 & Ridge CLIP (Polaris)$^{\dagger}$ & 61.41 & $-$1.70 [$-$2.52, $-$0.92] & 0.0 & $+$1.30 [$+$1.09, $+$1.50] \\
 & Ridge CLIP (Spica)$^{\dagger}$ & 59.27 & $-$3.84 [$-$4.78, $-$2.93] & 0.0 & $+$0.79 [$+$0.63, $+$0.97] \\
\addlinespace
Nebula test & Polos (official)$^{\dagger}$ & 53.67 & $-$2.20 [$-$3.30, $-$1.11] & 0.0 & -- \\
 & Polos$^{\dagger}$ & 54.31 & $-$1.56 [$-$2.65, $-$0.46] & 0.2 & $+$1.69 [$+$1.17, $+$2.22] \\
 & Rigel$^{\dagger}$ & 55.02 & $-$0.85 [$-$2.08, $+$0.43] & 8.8 & -- \\
 & Rigel +\,ref$^{\dagger}$ & 57.52 & $+$1.65 [$+$0.87, $+$2.44] & 100.0 & -- \\
 & DENEB$^{\ddagger}$ & 53.44 & $-$2.43 [$-$3.44, $-$1.43] & 0.0 & $-$0.58 [$-$0.82, $-$0.34] \\
 & Pearl$^{\dagger}$ & 54.91 & $-$0.96 [$-$1.96, $+$0.04] & 2.7 & $+$0.66 [$+$0.20, $+$1.12] \\
 & Ridge SigLIP (Polaris)$^{\dagger}$ & 55.55 & $-$0.32 [$-$1.28, $+$0.66] & 25.6 & $+$0.86 [$+$0.69, $+$1.04] \\
 & Ridge SigLIP (Spica)$^{\dagger}$ & 54.46 & $-$1.40 [$-$2.51, $-$0.32] & 0.2 & $+$0.63 [$+$0.43, $+$0.83] \\
 & Ridge CLIP (Polaris)$^{\dagger}$ & 53.63 & $-$2.24 [$-$3.34, $-$1.15] & 0.0 & $+$0.80 [$+$0.59, $+$1.01] \\
 & Ridge CLIP (Spica)$^{\dagger}$ & 52.69 & $-$3.18 [$-$4.39, $-$1.99] & 0.0 & $+$0.65 [$+$0.43, $+$0.89] \\
\addlinespace
Flickr8k-Expert & Polos$^{\ddagger}$ & 54.89 & $-$4.98 [$-$5.67, $-$4.31] & 0.0 & $+$0.23 [$-$0.01, $+$0.47] \\
 & DENEB$^{\ddagger}$ & 56.07 & $-$3.81 [$-$4.46, $-$3.17] & 0.0 & $-$0.28 [$-$0.38, $-$0.17] \\
 & Pearl$^{\ddagger}$ & 57.88 & $-$1.99 [$-$2.55, $-$1.43] & 0.0 & $-$0.60 [$-$0.82, $-$0.38] \\
 & Ridge SigLIP (Polaris)$^{\ddagger}$ & 58.32 & $-$1.55 [$-$2.10, $-$0.99] & 0.0 & $+$0.25 [$+$0.17, $+$0.33] \\
 & Ridge SigLIP (Spica)$^{\ddagger}$ & 58.90 & $-$0.97 [$-$1.50, $-$0.43] & 0.0 & $+$0.45 [$+$0.34, $+$0.56] \\
 & Ridge CLIP (Polaris)$^{\ddagger}$ & 57.15 & $-$2.73 [$-$3.33, $-$2.11] & 0.0 & $+$0.18 [$+$0.09, $+$0.28] \\
 & Ridge CLIP (Spica)$^{\ddagger}$ & 57.68 & $-$2.20 [$-$2.78, $-$1.60] & 0.0 & $+$0.29 [$+$0.17, $+$0.41] \\
\addlinespace
Composite & Polos$^{\ddagger}$ & 58.04 & $-$8.25 [$-$8.91, $-$7.59] & 0.0 & $-$0.77 [$-$0.99, $-$0.55] \\
 & DENEB$^{\ddagger}$ & 59.15 & $-$7.14 [$-$7.73, $-$6.56] & 0.0 & $-$0.43 [$-$0.55, $-$0.32] \\
 & Pearl$^{\ddagger}$ & 61.18 & $-$5.10 [$-$5.64, $-$4.57] & 0.0 & $-$1.11 [$-$1.29, $-$0.93] \\
 & Ridge SigLIP (Polaris)$^{\ddagger}$ & 59.26 & $-$7.03 [$-$7.62, $-$6.44] & 0.0 & $-$0.13 [$-$0.19, $-$0.07] \\
 & Ridge SigLIP (Spica)$^{\ddagger}$ & 62.16 & $-$4.12 [$-$4.64, $-$3.59] & 0.0 & $+$0.07 [$-$0.03, $+$0.16] \\
 & Ridge CLIP (Polaris)$^{\ddagger}$ & 56.68 & $-$9.61 [$-$10.26, $-$8.96] & 0.0 & $-$0.52 [$-$0.60, $-$0.44] \\
 & Ridge CLIP (Spica)$^{\ddagger}$ & 59.23 & $-$7.05 [$-$7.67, $-$6.45] & 0.0 & $-$0.11 [$-$0.21, $-$0.00] \\
\addlinespace
Flickr8k-CF & Polos$^{\ddagger}$ & 37.25 & $-$3.17 [$-$3.50, $-$2.85] & 0.0 & $+$0.25 [$+$0.15, $+$0.35] \\
 & DENEB$^{\ddagger}$ & 37.80 & $-$2.61 [$-$2.95, $-$2.29] & 0.0 & $-$0.27 [$-$0.32, $-$0.21] \\
 & Pearl$^{\ddagger}$ & 38.68 & $-$1.73 [$-$2.01, $-$1.46] & 0.0 & $-$0.13 [$-$0.20, $-$0.06] \\
 & Ridge SigLIP (Polaris)$^{\ddagger}$ & 39.03 & $-$1.39 [$-$1.66, $-$1.12] & 0.0 & $+$0.02 [$-$0.01, $+$0.05] \\
 & Ridge SigLIP (Spica)$^{\ddagger}$ & 39.69 & $-$0.72 [$-$0.98, $-$0.47] & 0.0 & $-$0.01 [$-$0.06, $+$0.05] \\
 & Ridge CLIP (Polaris)$^{\ddagger}$ & 37.70 & $-$2.72 [$-$3.03, $-$2.41] & 0.0 & $-$0.02 [$-$0.07, $+$0.03] \\
 & Ridge CLIP (Spica)$^{\ddagger}$ & 38.36 & $-$2.05 [$-$2.36, $-$1.76] & 0.0 & $+$0.01 [$-$0.05, $+$0.07] \\
\bottomrule
\end{tabular}
}\end{table}

  \begin{table}[!htbp]\caption{The six metrics closest to or above the human reference above the lowest score (paired difference on the same samples, uncorrected 95\% interval).}\label{tab:human}\centering\apptablefont\makebox[\linewidth]{\begin{tabular}{lrr}
\toprule
Metric & $\tau$ & Metric $-$ human [95\% CI] \\
\midrule
\multicolumn{3}{l}{\emph{Polaris test} (19,909 samples; human 31.06)} \\
\quad Judge InternVL-8B +\,ref & 33.47 & $+$2.40 [$+$1.27, $+$3.54] \\
\quad Judge Qwen-8B +\,ref & 32.86 & $+$1.80 [$+$0.67, $+$2.96] \\
\quad Judge Qwen-4B +\,ref & 32.81 & $+$1.75 [$+$0.60, $+$2.92] \\
\quad Ridge SigLIP (Polaris) & 31.43 & $+$0.37 [$-$0.88, $+$1.64] \\
\quad Judge Qwen-4B & 30.66 & $-$0.40 [$-$1.72, $+$0.98] \\
\quad Polos & 30.49 & $-$0.58 [$-$1.88, $+$0.78] \\
\addlinespace
\multicolumn{3}{l}{\emph{Spica test} (10,063 samples; human 30.17)} \\
\quad Judge InternVL-8B +\,ref & 33.90 & $+$3.73 [$+$2.21, $+$5.32] \\
\quad Judge Qwen-8B +\,ref & 33.21 & $+$3.04 [$+$1.49, $+$4.67] \\
\quad Judge Qwen-4B +\,ref & 33.14 & $+$2.97 [$+$1.40, $+$4.62] \\
\quad Judge Qwen-4B & 31.73 & $+$1.56 [$-$0.18, $+$3.38] \\
\quad Ridge SigLIP (Polaris) & 31.02 & $+$0.85 [$-$1.02, $+$2.71] \\
\quad DENEB & 30.80 & $+$0.63 [$-$1.18, $+$2.51] \\
\addlinespace
\multicolumn{3}{l}{\emph{Flickr8k-Expert} (7,196 samples; human 52.52)} \\
\quad Rigel & 53.32 & $+$0.79 [$-$0.33, $+$1.98] \\
\quad Rigel +\,ref & 53.01 & $+$0.49 [$-$0.63, $+$1.67] \\
\quad Ridge SigLIP (Spica) & 52.89 & $+$0.37 [$-$0.77, $+$1.53] \\
\quad Judge InternVL-8B +\,ref & 52.71 & $+$0.19 [$-$0.98, $+$1.38] \\
\quad Judge Qwen-8B +\,ref & 52.26 & $-$0.26 [$-$1.49, $+$0.98] \\
\quad Judge Qwen-8B & 52.21 & $-$0.31 [$-$1.51, $+$0.90] \\
\bottomrule
\end{tabular}
}\end{table}

  \begin{table}[!htbp]\caption{Paired differences between the learned regression metrics (points of $\tau$). Asterisk: significant after Holm correction over all pairs of the set.}\label{tab:learned}\centering\apptablefont\makebox[\linewidth]{\begin{tabular}{llrrr}
\toprule
Evaluation set & Samples & Polos $-$ DENEB & Polos $-$ Pearl & DENEB $-$ Pearl \\
\midrule
Polaris test & all & $+$0.62 & $+$1.28$^{*}$ & $+$0.66 \\
 & above lowest & $+$0.72 & $+$2.69$^{*}$ & $+$1.98$^{*}$ \\
Spica test & all & $-$0.94 & $-$0.44 & $+$0.50 \\
 & above lowest & $-$1.51 & $-$0.09 & $+$1.42 \\
Nebula test & all & $-$1.41 & $-$1.64 & $-$0.23 \\
 & above lowest & $-$2.22 & $-$2.65 & $-$0.44 \\
Flickr8k-Expert & all & $-$1.68$^{*}$ & $-$3.81$^{*}$ & $-$2.14$^{*}$ \\
 & above lowest & $-$1.88 & $-$3.33$^{*}$ & $-$1.46 \\
Composite & all & $-$0.78 & $-$3.49$^{*}$ & $-$2.71$^{*}$ \\
 & above lowest & $-$0.46 & $-$2.79$^{*}$ & $-$2.33$^{*}$ \\
Flickr8k-CF & all & $-$1.07$^{*}$ & $-$1.81$^{*}$ & $-$0.74$^{*}$ \\
 & above lowest & $-$4.85$^{*}$ & $-$4.05$^{*}$ & $+$0.79 \\
\bottomrule
\end{tabular}
}\end{table}

  \begin{table}[!htbp]\caption{How well each set separates $n$ ranked metrics. \%: share of all metric pairs separated by a paired bootstrap test (Holm-corrected, $\alpha=0.05$). Top 8: separated pairs among the 28 pairs of the eight best metrics. Tied: metrics not separated from the first.}\label{tab:power}\centering\apptablefont\makebox[\linewidth]{\begin{tabular}{lrrrrrrrrrr}
\toprule
& & \multicolumn{3}{c}{$\tau$, all samples} & \multicolumn{3}{c}{$\tau$, above lowest} & \multicolumn{3}{c}{Order accuracy} \\
\cmidrule(lr){3-5} \cmidrule(lr){6-8} \cmidrule(l){9-11}
Evaluation set & $n$ & \% & top 8 & tied & \% & top 8 & tied & \% & top 8 & tied \\
\midrule
Polaris test & 31 & 82 & 12 & 3 & 73 & 13 & 3 & 18 & 0 & 17 \\
Spica test & 30 & 79 & 10 & 3 & 66 & 10 & 3 & 7 & 0 & 18 \\
Nebula test & 30 & 72 & 1 & 7 & 57 & 1 & 8 & -- & -- & -- \\
\addlinespace
Flickr8k-Expert & 33 & 83 & 15 & 1 & 70 & 2 & 8 & 66 & 9 & 4 \\
Composite & 33 & 90 & 20 & 2 & 84 & 15 & 3 & -- & -- & -- \\
Flickr8k-CF & 33 & 90 & 12 & 3 & 80 & 16 & 2 & -- & -- & -- \\
\bottomrule
\end{tabular}
}\end{table}

\section{Further case studies}
\label{app:cases}

Figs.~\ref{fig:casesone} and~\ref{fig:casestwo} show twelve further pairs in the layout of Fig.~\ref{fig:cases}: seven from the Polaris test set, three from Flickr8k-Expert and two from Composite. They were selected from the largest disagreements between families of metrics and people, and from the closest agreements, so that each pattern we met repeatedly is represented. Rigel is shown on Flickr8k-Expert and Composite, where it is ranked.

\emph{Agreement at both ends.} A caption that belongs to another image (Fig.~\ref{fig:casestwo}a) is placed at the bottom by every metric, and a correct, specific caption (Fig.~\ref{fig:casestwo}d) at the top by all but CIDEr. Cases like the first are what the detection term of Eq.~\eqref{eq:components} rewards.

\emph{Different wording.} Text overlap underrates captions that are right but phrased unlike the references, or that contain a misspelt word (Figs.~\ref{fig:casesone}a, \ref{fig:casesone}c, \ref{fig:casestwo}d).

\emph{One wrong detail.} When a fluent caption misnames the animal, the action or the number of people, embedding and learned regression metrics tend to place it high. The judge is closest to people in some of these cases (Figs.~\ref{fig:casesone}b, \ref{fig:casestwo}b, \ref{fig:casestwo}f) and as wrong as the others, or more so, in others (Figs.~\ref{fig:casesone}d, \ref{fig:casestwo}e).

\emph{Failures of the judge.} The judge underrates a correct caption with a misspelt word (Fig.~\ref{fig:casesone}c) and overrates a plausible caption for a scene it does not describe (Fig.~\ref{fig:casestwo}c), where every other metric agrees with people.

\emph{Debatable human scores.} In Fig.~\ref{fig:casesone}e people rate a caption about a minor part of the image highly and all metrics place it at the bottom; in Fig.~\ref{fig:casesone}f people give a middling score to a caption that the judge places at the top. Such cases limit the agreement that any metric can reach and are one reason why the human reference of Eq.~\eqref{eq:human} is not a ceiling.

The percentile scale hides how far apart two pairs are in absolute score, so a case is evidence for the direction of an error and not for its size.

\begin{figure}[!htbp]
\centering
\includegraphics[width=\linewidth]{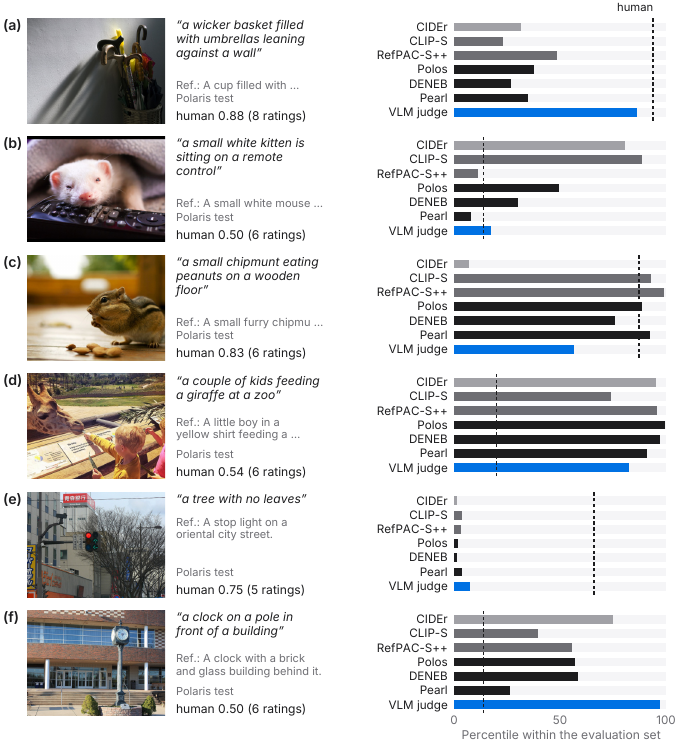}
\caption{Further case studies (1 of 2). Layout as in Fig.~\ref{fig:cases}; the grey line above the human score names the evaluation set. All six pairs are from the Polaris test set.}
\label{fig:casesone}
\end{figure}

\begin{figure}[!htbp]
\centering
\includegraphics[width=\linewidth]{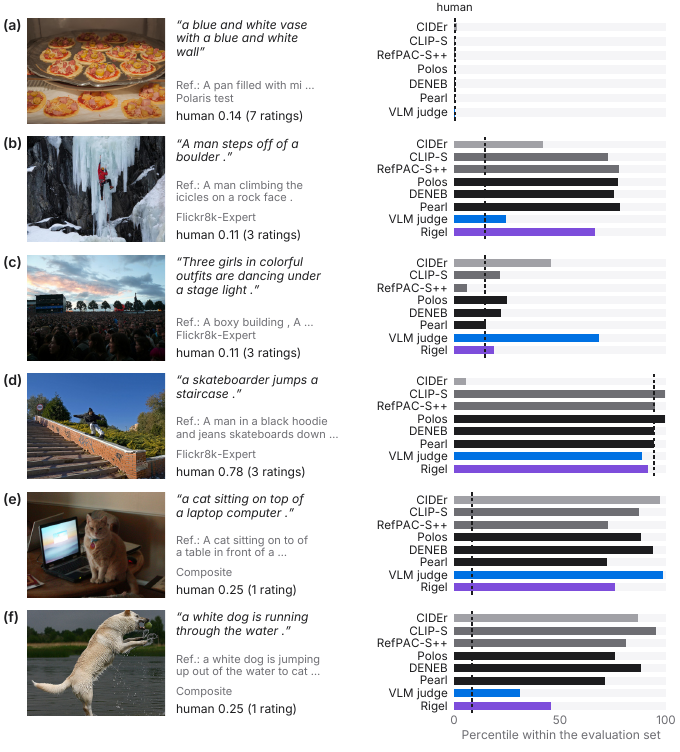}
\caption{Further case studies (2 of 2): one pair from the Polaris test set, three from Flickr8k-Expert and two from Composite.}
\label{fig:casestwo}
\end{figure}
\fi

\ifdefined\WithoutAppendix\else\clearpage\fi
\bibliographystyle{elsarticle-num}
\bibliography{refs}

\end{document}